%% file: sample-sigconf.tex
\documentclass[sigconf]{acmart}
\usepackage{hyperref}
\usepackage{colortbl}
\usepackage{float}

\usepackage{graphicx}
\usepackage{xspace}
\usepackage[table]{xcolor}
\usepackage{booktabs} 
\usepackage{multirow}
\usepackage{array}
\usepackage{amsmath}
\usepackage{tcolorbox}
\usepackage{bbm}

\newcommand{\eg}{\emph{e.g.,}\xspace}
\newcommand{\ie}{\emph{i.e.,}\xspace}

\newcommand{\wrt}{\emph{w.r.t}\xspace}

\definecolor{lightgrayv}{HTML}{F4F3F8} 
\definecolor{mydarkblue}{rgb}{0,0.08,0.45}
\newcommand{\baby}{\textsc{LonsRex}\xspace}

\AtBeginDocument{%
  }

\setcopyright{acmlicensed}
\copyrightyear{2026}
\acmYear{2026}
\acmDOI{XXXXXXX.XXXXXXX}

\acmConference[KDD '26]{the 32st ACM SIGKDD Conference on Knowledge Discovery and Data Mining}{9–13 August, 2026}{Jeju, Korea}
\acmISBN{978-1-4503-XXXX-X/18/06}

\begin{document}

\title{Are Rationales Necessary and Sufficient? Tuning LLMs for Explainable Misinformation Detection}

\def\huggingface{\raisebox{-1.5pt}{\includegraphics[height=1.0em]{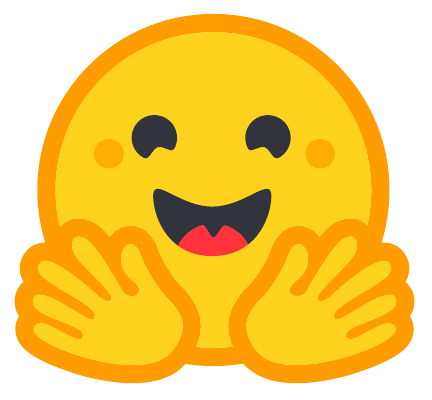}}}
\def\github{\raisebox{-1.5pt}{\includegraphics[height=1.0em]{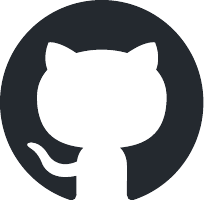}}}

\author{Bing Wang}
\orcid{0000-0002-1304-3718}
\affiliation{
  \institution{College of Computer Science and Technology, Jilin University}
  \city{Changchun}
  \country{China}}
\affiliation{
    \institution{Tongyi Lab, Alibaba Group}
    \city{Hangzhou}
    \country{China}}
\email{wangbing1416@gmail.com}

\author{Rui Miao}
\orcid{0000-0002-2917-2311}
\affiliation{
  \institution{School of Artificial Intelligence, Jilin University}
  \city{Changchun}
  \country{China}}
\email{miaorui24@mails.jlu.edu.cn}

\author{Ximing Li}
\authornote{Ximing Li and Chen Shen are the corresponding authors. Bing Wang, Ximing Li, Changchun Li, and Xiaosong Yuan are also affiliated with Key Laboratory of Symbolic Computation and Knowledge Engineering of the Ministry of Education, Jilin University.}
\orcid{0000-0001-8190-5087}
\affiliation{
  \institution{College of Computer Science and Technology, Jilin University}
  \city{Changchun}
  \country{China}}
\affiliation{
  \institution{RIKEN AIP}
  \city{Tokyo}
  \country{Japan}}
\email{liximing86@gmail.com}

\author{Chen Shen}
\authornotemark[1]
\orcid{0000-0002-7534-0830}
\author{Shaotian Yan}
\orcid{0000-0002-9029-0763}
\affiliation{
    \institution{Tongyi Lab, Alibaba Group}
    \city{Hangzhou}
    \country{China}}
\email{zjushenchen@gmail.com}
\email{yanshaotian@gmail.com}

\author{Changchun Li}
\orcid{0000-0002-8001-2655}
\affiliation{
  \institution{College of Computer Science and Technology, Jilin University}
  \city{Changchun}
  \country{China}}
\email{changchunli93@gmail.com}

\author{Kaiyuan Liu}
\orcid{0009-0005-4717-0087}
\affiliation{
    \institution{College of Computer Science and Technology, Zhejiang University}
    \city{Hangzhou}
    \country{China}}
\email{12421281@zju.edu.cn}

\author{Xiaosong Yuan}
\orcid{0000-0001-5748-5174}
\affiliation{
    \institution{College of Computer Science and Technology, Jilin University}
    \city{Changchun}
    \country{China}}
\email{yuanxs19@mails.jlu.edu.cn}

\author{Jieping Ye}
\orcid{0000-0001-8662-5818}
\affiliation{
    \institution{Tongyi Lab, Alibaba Group}
    \city{Hangzhou}
    \country{China}}
\email{yejieping.ye@alibaba-inc.com}

\renewcommand{\shortauthors}{Wang et al.}

\begin{abstract}
  \input{S_Abstract}
\end{abstract}

\ccsdesc[500]{Computing methodologies~Artificial intelligence}
\ccsdesc[500]{Information systems~Social networks}

\keywords{Social media, misinformation detection, large language models, explainability, supervised fine-tuning, data filtering}


\maketitle

\input{S_Introduction}

\input{S_Preliminary}

\input{S_Method}

\input{S_Experiment}

\input{S_RelatedWorks}

\input{S_Conclusion}

\section*{Acknowledgement}

This work was supported by the National Natural Science Foundation of China (No.62276113) and Alibaba Research Intern Program.

\clearpage
\bibliographystyle{ACM-Reference-Format}
\bibliography{reference}

\input{S_Appendix}


\end{document}

%% file: S_Abstract.tex
The rapid spread of misinformation on social media platforms has become a formidable challenge. To mitigate its proliferation, Misinformation Detection (MD) has emerged as a critical research topic. Traditional MD approaches based on small models typically perform binary classification, \eg real and fake, through a black-box process. Recently, the rise of Large Language Models (LLMs) has enabled \textbf{explainable MD}, where models generate rationales that explain their decisions, thereby enhancing transparency. 
Existing explainable MD methods primarily focus on crafting sophisticated prompts to elicit rationales from off-the-shelf LLMs.
In this work, we propose a pipeline to fine-tune a dedicated LLM specifically for explainable MD. Our pipeline begins by collecting large-scale fact-checked articles, and then uses multiple strong LLMs to produce veracity predictions and rationales. To ensure high-quality training data, we leverage a filtering strategy that selects only the correct instances for fine-tuning. 
While this pipeline is intuitive and prevalent, our experiments reveal that naive filtering based solely on label correctness is insufficient in practice and suffers from two critical limitations:
(1) \textbf{Coarse-grained labels cause insufficient rationales}: 
Rationales filtered solely based on binary labels are insufficient to adequately support their decisions;
(2) \textbf{Over-verification behavior causes unnecessary rationales}: Stronger LLMs tend to exhibit over-verification behavior, producing excessively verbose and unnecessary rationales.
To address these issues, we introduce \textbf{\baby}, a novel data synthesis pipeline to \textbf{Lo}cate \textbf{N}ecessary and \textbf{S}ufficient \textbf{R}ationales for \textbf{Ex}plainable MD. Specifically, we propose a metric that quantifies the contribution of each verification step to the final prediction, thereby evaluating its necessity and sufficiency. Experimental results demonstrate that \baby improves the accuracy of baseline LLMs by approximately 22.97\% and is comparable to larger LLMs. We will publicly release our 316k raw data and the filtered version by \baby.

\begin{center}
\noindent
\huggingface \quad \href{https://huggingface.co/datasets/wangbing1416/LonsRex-Misinfomation-SFT}{\color{mydarkblue} \textit{\textbf{Data and LLMs}}} \quad
\github \quad \href{https://github.com/wangbing1416/LONSREX}{\color{mydarkblue} \textbf{\textit{Code}}}
\end{center}

%% file: S_Introduction.tex
\section{Introduction}

On today’s social media platforms, \eg X.com and Reddit, vast numbers of users actively disseminate and consume information. Unfortunately, this information ecosystem is inevitably polluted by a substantial amount of deliberately fabricated misinformation, which poses serious threats to users’ financial security and psychological well-being \citep{lazer2018science,vosoughi2018spread}. To mitigate the spread of such misinformation content, Misinformation Detection (MD) has emerged as a critical and active research topic, aiming to automatically and rapidly identify misinformation on social media to facilitate timely intervention and control \citep{hu2024bad,wang2025collaboration,modzelewski2025pcot}.

\begin{figure}[t]
  \centering
  \includegraphics[width=0.88\columnwidth]{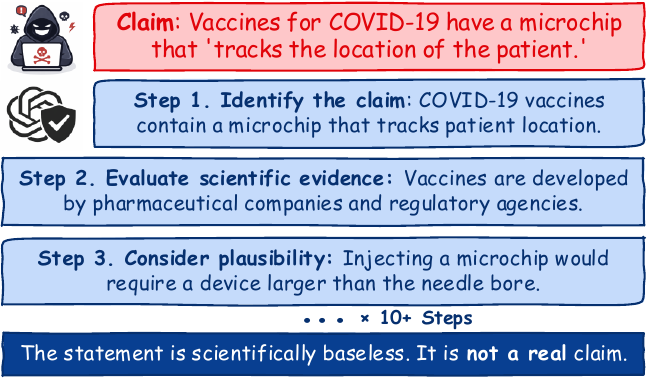}
  \caption{A representative case demonstrate one rationale generated by LLMs for fake claims.}
  \label{intro}
  \vspace{-9pt}
\end{figure}

Typically, prior MD methods involve training discriminative models to classify claims on social media into binary labels, \ie real or fake, by incorporating various external features, \eg emotion signals \citep{zhang2021mining,liu2025raemollm}, commonsense knowledge \citep{dun2021kan,wang2025robust}, and publisher intent \citep{wang2024why,wang2025exploring}. 
Despite their superior performance, the black-box nature of these discriminative models prevents them from revealing the underlying reasoning behind their predictions. Accordingly, to improve decision transparency, recent research has shifted toward \textbf{explainable MD}, which seeks to generate \textbf{natural language rationales} in addition to the binary output \citep{yang2022a,wang2024explainable,wang2025llm}.
They primarily leverage Large Language Models (LLMs) to design sophisticated prompts to directly elicit rationales \citep{hu2024bad,modzelewski2025pcot} or summarize retrieved external evidence to serve as rationales \citep{wang2024explainable,yue2024evidence}. For example, PCoT \citep{modzelewski2025pcot} employs a system prompt grounded in persuasion knowledge; L-Defense \citep{wang2024explainable} clusters retrieved evidence into supporting and opposing groups and uses an LLM to summarize them, respectively.

Although these approaches outperform straightforward zero-shot prompting with LLMs, they heavily depend on the capabilities of the foundation LLMs and the quality of the retrieved evidence, which limits their potential. In this work, we take a different perspective and design a data-centric pipeline that \textbf{fine-tunes a lightweight LLM specifically for explainable MD}.
Specifically, our pipeline begins by collecting 316k publicly available fact-checked claims, each labeled as either real or fake. We then prompt multiple strong LLMs to generate corresponding step-by-step rationales for these claims, as illustrated in Fig.~\ref{intro}. Finally, some heuristic filtering criteria, \eg correctness \citep{chen2025skip,luo2025deconstructing} and formatting standards \citep{guo2025deepseek}, are applied to select higher-quality rationale-label pairs for supervised fine-tuning of a lightweight LLM, \eg Qwen3-4B-Instruct \citep{yang2025qwen3}.

While this pipeline is intuitive and prevalent, our experiments further reveal that naive filtering based solely on binary label correctness is insufficient in practice for explainable MD and suffers from two critical limitations: 
\textbf{(1) Coarse-grained labels cause insufficient rationales}: As illustrated in Fig.~\ref{intro}, the rationale \wrt a claim typically involves more than ten verification steps. Relying solely on the correctness of binary labels to filter training samples is insufficient to evaluate the quality of each verification step. Therefore, the training data inevitably contain rationales that are meaningless or even noisy, which is empirically analyzed in Sec.~\ref{sec2.2}.
\textbf{(2) Over-verification behavior causes unnecessary rationales}: Our experiments in Sec.~\ref{sec2.3} demonstrate that LLMs with stronger reasoning capabilities tend to generate more complex verification processes, leading to overly strict claim verification. Such over-verification not only increases the likelihood of labeling claims as fake, but also degrades the performance of LLMs trained on these rationales, as excessive and unnecessary verification steps introduce noise and reduce training efficiency.

To alleviate these issues, we update our training pipeline and propose a new method, namely \textbf{Lo}cating \textbf{N}ecessary and \textbf{S}ufficient \textbf{R}ationales for \textbf{Ex}plainable MD (\textbf{\baby}). 
Specifically, the basic idea of \baby is to design a perplexity-based metric that quantifies the contribution of each verification step within rationales to the prediction, and to select necessary and sufficient rationales that contain fewer useless verification steps and more useful ones.
To achieve this, we first evaluate the contribution of each verification step by measuring the change in the LLM’s prediction perplexity when that step is masked. Building on this counterfactual attribution framework, we introduce two complementary attribution metrics: \textbf{self-attribution} and \textbf{mutual-attribution} scores. 
The self-attribution score evaluates the average contribution of individual verification steps within a single rationale; The mutual-attribution score summarizes diverse verification perspectives across multiple rationales, and assumes that frequently occurring high-contribution perspectives are generally more important; accordingly, it assesses the average contribution of perspectives within each rationale.
During the implementation, we generate 948k rationales and predictions, and select 200k high-quality ones to fine-tune LLMs of varying sizes, \eg 4B, 2B, and 1.5B parameters. The resulting LLMs significantly outperform their foundation counterparts and achieve performance comparable to that of much larger open-source LLMs.

Our contributions can be summarized as the following three-fold:
\begin{itemize}
    \item We empirically identify that rationale-based training data for explainable MD, generated by LLMs, often contains an amount of unnecessary and insufficient rationales.  
    \item To address this, we propose a novel pipeline \baby that fine-tunes LLMs to generate high-quality MD rationales by explicitly locating those necessary and sufficient rationales.
    \item We further construct an evaluation benchmark for explainable MD and demonstrate that our fine-tuned LLMs consistently outperform their foundation LLMs and achieve competitive results against much larger open-source LLMs.
\end{itemize}

%% file: S_Preliminary.tex
\vspace{-3pt}
\section{Preliminary Empirical Analysis} \label{sec2}

We empirically evaluate the training rationales generated by advanced LLMs and find that they contain numerous unnecessary and insufficient rationales, even when their predictions are correct.

\noindent
\textbf{Task formulation of explainable MD.}
Formally, given a claim $\mathbf{x}$ from social media platforms, explainable MD aims to predict a veracity label $y \in \{0,1\}$ (0/1 means fake/real) and generate a corresponding rationale $\mathbf{E} = \{\mathbf{e}_{1}, \ldots, \mathbf{e}_{L}\}$. Here, $\mathbf{e}_{l}$ denotes the $l$-th verification step that analyzes the claim from a specific perspective, as illustrated in Fig.~\ref{intro}.
In this work, we collect a large-scale fact-checked dataset $\mathcal{D} = \{(\mathbf{x}_i, y_i)\}_{i=1}^{|\mathcal{D}|}$ from open-sourced repositories. We leverage advanced LLMs to generate a set of candidate rationales and predictions $\{(\mathbf{E}_{ik}, \widehat y_{ik})\}_{k=1}^{K}$ for each claim $\mathbf{x}_i$, where $K$ denotes the number of candidates per claim.
Then, we apply heuristic filtering strategies, such as filtering based on veracity labels $\widehat y$, to curate a high-quality subset $\mathcal{\widehat D} = \{(\mathbf{x}_i, \mathbf{E}_i, \widehat y_i)\}_{i=1}^{|\mathcal{\widehat D}|}$, where $|\mathcal{\widehat D}| < K |\mathcal{D}|$. Upon this refined subset, we fine-tune a lightweight LLM, parameterized by $\boldsymbol{\theta}$, by minimizing the negative log-likelihood loss as follows:
\begin{equation}
    \label{eq1}
    \mathop{\boldsymbol{\min}}_{\boldsymbol{\theta}}  \mathcal{L}(\boldsymbol{\theta}) = - \frac{1}{|\mathcal{\widehat D}|} \sum \nolimits_{(\mathbf{x}_i, \mathbf{E}_i, \widehat y_i) \in \mathcal{\widehat D}} \log P_{\boldsymbol{\theta}} \left(\mathbf{E}_i, \widehat y_i \mid \mathbf{x}_i \right).
\end{equation}

\noindent
\textbf{Evaluation settings.}
Our work aims to evaluate the necessity and sufficiency of rationales by measuring the contribution of each verification step to the veracity prediction. We hypothesize that a rationale is \textit{unnecessary} if it contains a high proportion of useless steps; conversely, it is deemed \textit{insufficient} if it contains a relatively low ratio of useful steps.
To quantify this, inspired by counterfactual attribution \citep{chen2023disco}, we measure the marginal impact of each step by observing the change in the model's output probability after it is masked. Formally, given a complete rationale $\mathbf{E} = \{\mathbf{e}_{l}\}_{l=1}^L$ and its perturbed version $\mathbf{E}_{\setminus l} = \mathbf{E} \setminus \{\mathbf{e}_l\}$ with the $l$-th step removed, the contribution $\Delta_l$ of step $\mathbf{e}_l$ to the prediction $\widehat y$ is defined as follows:
\begin{equation}
    \label{eq:attribution}
    \Delta_l = \log P_{\boldsymbol{\theta}} \left( \widehat y \mid \mathbf{x}, \mathbf{E} \right) - \log P_{\boldsymbol{\theta}}\left( \widehat y \mid \mathbf{x}, \mathbf{E}_{\setminus l} \right),
\end{equation}
where $P_{\boldsymbol{\theta}} \left( \widehat y \mid \cdot \right)$ denotes the confidence score assigned by the off-the-shelf LLM $\boldsymbol{\theta}$ to the prediction $\widehat y$. A higher $\Delta_l$ indicates that $\mathbf{e}_l$ is a critical piece of evidence, whereas a $\Delta_l \approx 0$ suggests that the step is redundant or even obstructive to the verification process.

\subsection{Unnecessary and Insufficient Rationales Even Though Correct Predictions} \label{sec2.2}

Because during the training rationale filtering process, using label correctness is a straightforward and naive approach. However, for MD, the labels consist solely of real and fake. Our experiments in this section reveal that samples filtered using such coarse-grained labels, even when the final predictions are correct, still contain a large number of unnecessary and insufficient rationales.

\begin{figure}[t]
  \centering
  \includegraphics[width=0.9\columnwidth]{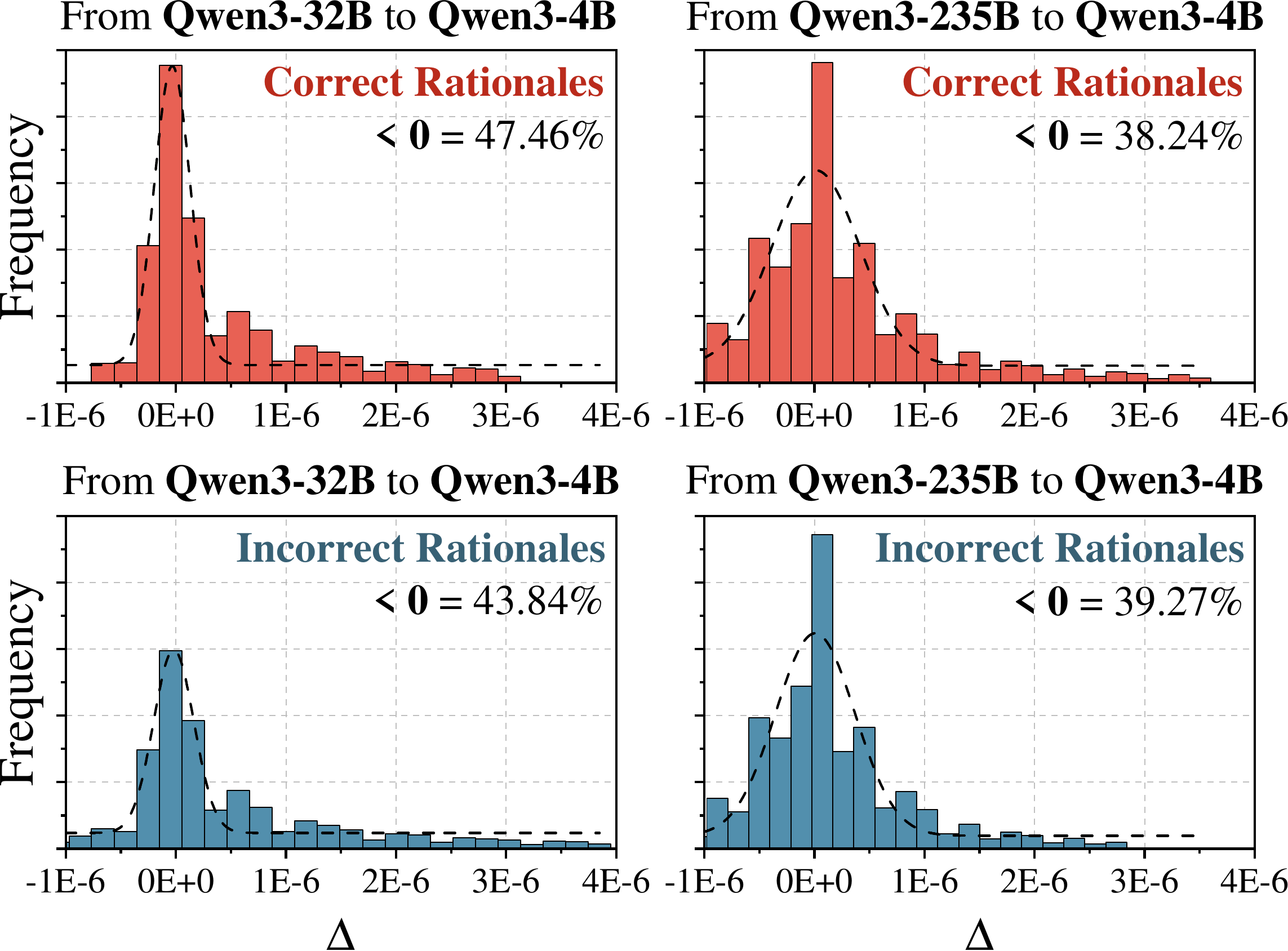}
  \vspace{-5pt}
  \caption{Distribution of $\Delta$ for correct / incorrect rationales.}
  \label{delta_correct_incorrect}
  \vspace{-5pt}
\end{figure}

\begin{figure}[t]
  \centering
  \includegraphics[width=0.95\columnwidth]{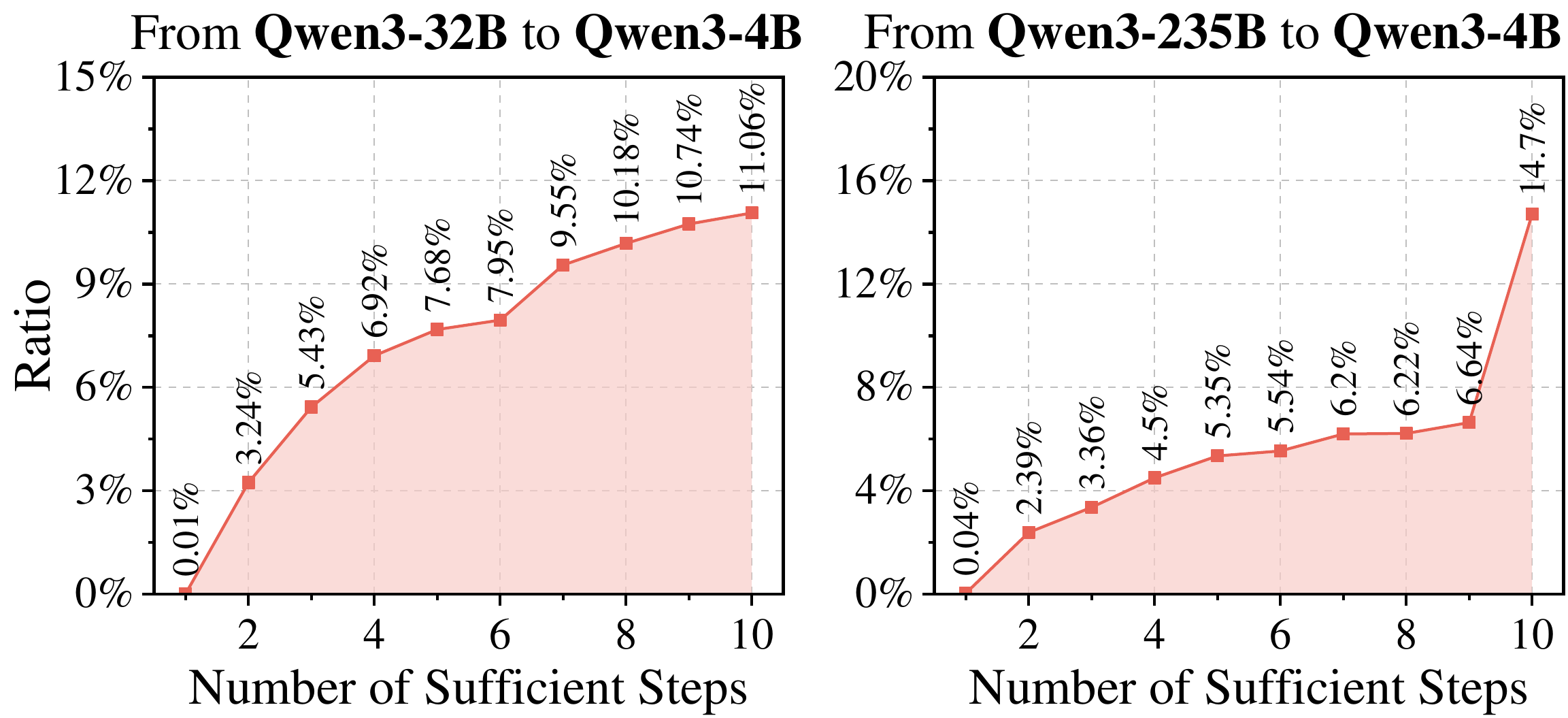}
  \vspace{-5pt}
  \caption{Distribution of the number of minimal sufficient verification steps $\kappa$ for correct rationales.}
  \label{statistic_sufficient}
  \vspace{-6pt}
\end{figure}

\begin{figure}[t]
  \centering
  \includegraphics[width=0.88\columnwidth]{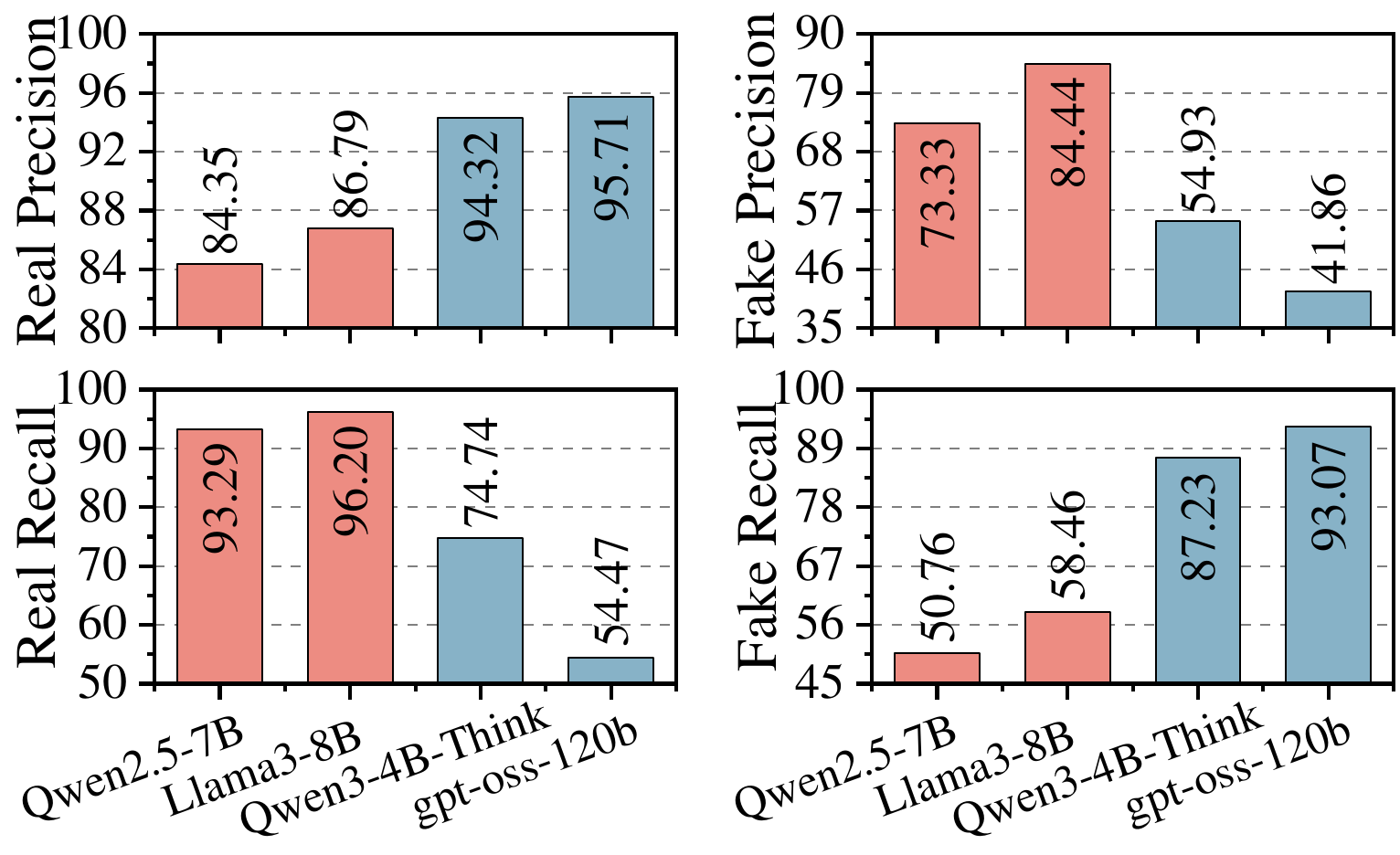}
  \vspace{-5pt}
  \caption{Precision and recall scores of instruct-tuned and reasoning LLMs on real and fake classes, respectively.}
  \label{think_performance}
  \vspace{-6pt}
\end{figure}

\vspace{2pt} \noindent
\textbf{Correct but unnecessary rationales. }
In Fig.~\ref{delta_correct_incorrect}, we illustrate the distribution of step contribution $\Delta$ for correct and incorrect rationales, based on data from Qwen3-32B and Qwen3-235B-A22B, during the fine-tuning of the Qwen3-4B-Instruct. The experimental results show that average 42.85\% of the steps in correct rationales have a $\Delta < 0$, meaning their contribution to the prediction is actually negative, and the majority of steps have a $\Delta$ concentrated around 0, indicating they provide no contribution to the prediction. Therefore, we can draw a preliminary experimental conclusion:
\vspace{-2pt}
\begin{tcolorbox}[
  colframe=black,
  boxrule=0.75pt,
  left=2.0pt, right=2.0pt, top=2.0pt, bottom=2.0pt, 
  fonttitle=\bfseries
]
\linespread{0.95}\selectfont
\textbf{$\boldsymbol{\star}$ Conclusion 1.}
Even when the \textbf{veracity predictions are correct}, there still exist a number of rationales containing useless, even negative, verification steps, \ie \textbf{unnecessary rationales}.
\end{tcolorbox}
\vspace{-2pt}
Furthermore, in Fig.~\ref{delta_correct_incorrect}, we also observe that in the incorrectly predicted samples, an average of 41.55\% of the steps have a contribution $\Delta < 0$. This indicates that these steps have a negative contribution to the incorrect prediction, meaning they actually provide a positive contribution toward the ground-truth label. Therefore, another empirical conclusion is that \textbf{even when predictions are incorrect, their rationales still contain partially useful steps}.

\vspace{2pt} \noindent
\textbf{Correct but insufficient rationales. }
To evaluate the sufficiency of the rationales, we rank the verification steps within each rationale by their contributions $\{\Delta_l\}_{l=1}^L$. We then select the top-$\kappa$ steps and compute the label's prediction probability using only these steps. A rationale is deemed sufficient if this restricted probability meets a predefined threshold $\epsilon = 0.01$ relative to the original log-probability. Formally, sufficiency is achieved if
\begin{equation}
\label{eq3}
    \log P_{\boldsymbol{\theta}} \left( \widehat{y} \mid \mathbf{x}, \mathbf{E}_{\text{top-}\kappa} \right) \geq (1 - \epsilon) \log P_{\boldsymbol{\theta}} \left( \widehat{y} \mid \mathbf{x}, \mathbf{E} \right),
\end{equation}
where $\mathbf{E}_{\text{top-}\kappa}$ denotes the set of the $\kappa$ highest-contributing steps. We then quantify the sufficiency of each rationale by calculating the minimum $\kappa$ required to satisfy this condition. The resulting distribution of $\kappa$ values of correct rationales is illustrated in Fig. \ref{statistic_sufficient}.

In practice, when humans judge misinformation, they often only require a very small number of key pieces of evidence to determine its falsity. However, Fig.~\ref{statistic_sufficient} shows that rationales achieving sufficiency with only two steps account for approximately 2.8\%, and those achieving sufficiency within four steps account for only about 12.9\%. Therefore, we can draw an experimental conclusion: 
\vspace{-2pt}
\begin{tcolorbox}[
  colframe=black,
  boxrule=0.75pt,
  left=2.0pt, right=2.0pt, top=2.0pt, bottom=2.0pt, 
  fonttitle=\bfseries
]
\linespread{0.95}\selectfont
\textbf{$\boldsymbol{\star}$ Conclusion 2.}
Even when the \textbf{veracity predictions are correct}, the proportion of \textbf{rationales that are sufficient} (containing useful steps) \textbf{is very small}.
\end{tcolorbox}
\vspace{-3pt}

\subsection{Over-Verification of LLMs} \label{sec2.3}

In this section, we find that training data generated by advanced LLMs often suffers from the problem of over-verification, particularly with LLMs possessing stronger reasoning capabilities, \eg gpt-oss-120b \citep{openai2025gpt}. This over-verification issue further increases the proportion of unnecessary rationales.

\begin{figure}[t]
  \centering
  \includegraphics[width=0.93\columnwidth]{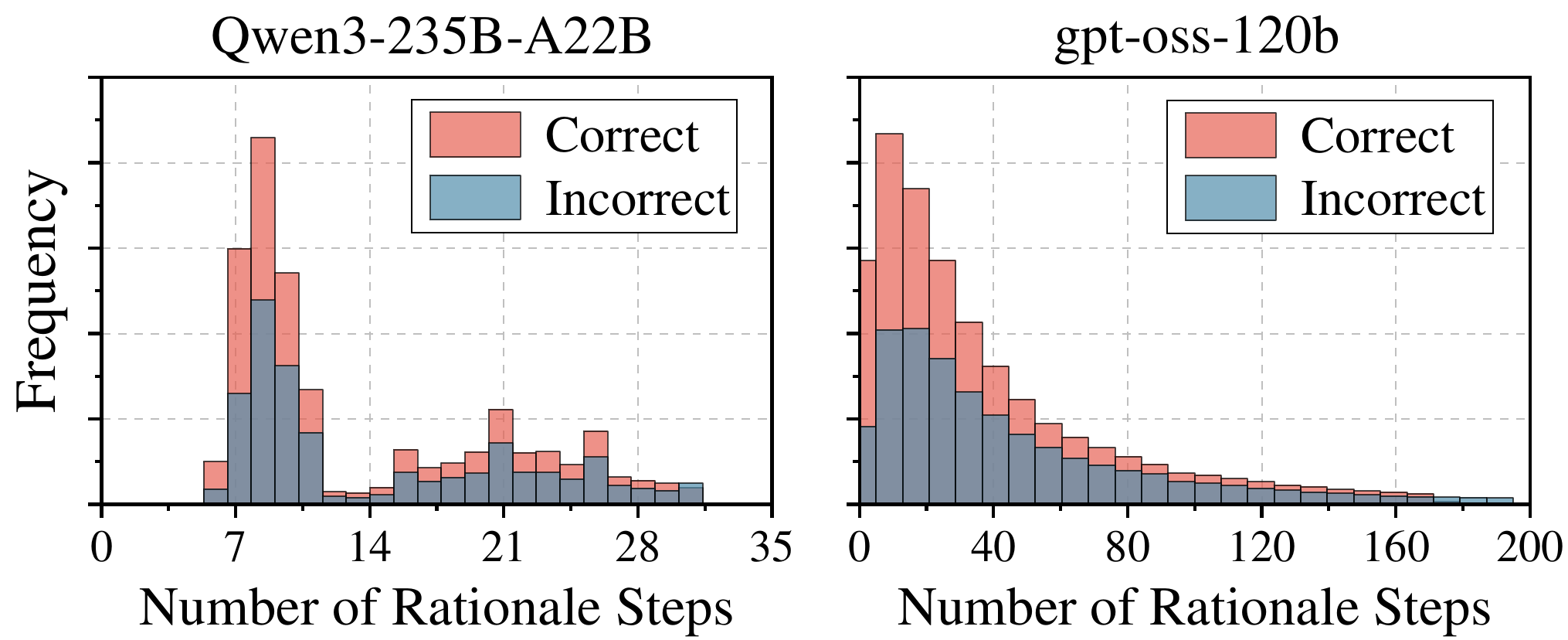}
  \vspace{-5pt}
  \caption{Distribution of the number of verification steps.}
  \label{statistic_step_number}
  \vspace{-8pt}
\end{figure}

\vspace{2pt} \noindent
\textbf{Over-verification behavior especially in strong reasoning LLMs. }
First, the results in Table~\ref{sota_results} demonstrate the MD performance of gpt-oss-120b, a strong open-source reasoning LLM, compared to several instruct-tuned LLMs of similar scale. The results reveal that this reasoning LLM consistently exhibits inferior detection performance. More specifically, in Fig.~\ref{think_performance}, we further compare the precision and recall of two instruct-tuned LLMs and two reasoning LLMs on the real and fake classes, respectively. It is clearly observed that reasoning LLMs achieve higher recall on the fake class, indicating a tendency for reasoning LLMs to classify information as fake more frequently.
To investigate the underlying cause of this phenomenon, we provide a representative case in Table~\ref{case}. We find that reasoning LLMs generally generate a significantly higher number of verification steps compared to instruct-tuned LLMs, a finding further corroborated by Fig.~\ref{statistic_step_number}. Therefore, if any single verification step concludes the information is fake, the LLM tends to predict the final label as fake. This abundance of verification steps leads the LLM to be biased toward the fake prediction, which is referred to as the \textbf{over-verification problem}.
These analyses induce a conclusion: 
\vspace{-2pt}
\begin{tcolorbox}[
  colframe=black,
  boxrule=0.75pt,
  left=2.0pt, right=2.0pt, top=2.0pt, bottom=2.0pt, 
  fonttitle=\bfseries
]
\linespread{0.95}\selectfont
\textbf{$\boldsymbol{\star}$ Conclusion 3.}
Stronger LLMs, particularly reasoning ones, always suffer from an over-verification problem in MD. Specifically, \textbf{their tendency to perform more verification steps causes their predictions to be biased toward the fake class}.
\end{tcolorbox}
\vspace{-2pt}

\begin{figure}[t]
  \centering
  \includegraphics[width=0.96\columnwidth]{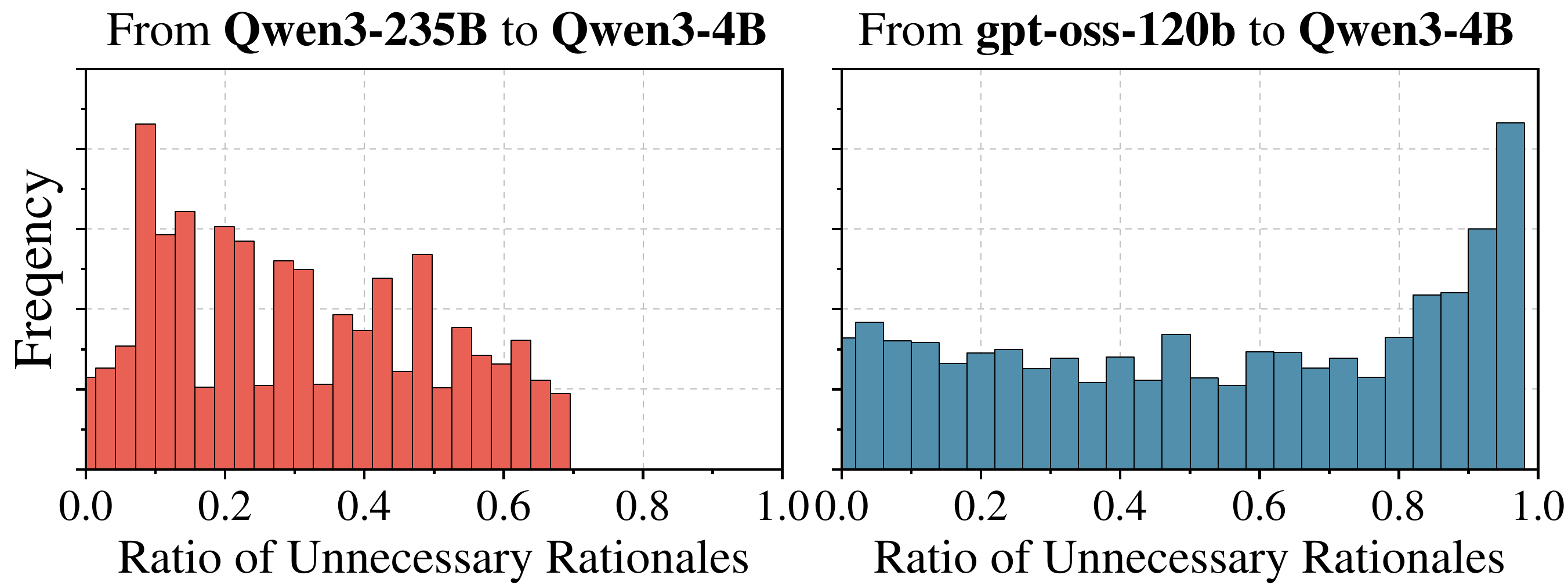}
  \vspace{-5pt}
  \caption{Distribution of the ratio of unnecessary rationales.}
  \label{unnecessary_ratio}
  \vspace{-8pt}
\end{figure}

\vspace{2pt} \noindent
\textbf{More unnecessary rationales due to over-verification. }
The over-verification problem leads to an increased number of verification steps, which intuitively results in more unnecessary rationales. To verify this, Fig.~\ref{unnecessary_ratio} shows the distribution of the ratio of unnecessary rationales for gpt-oss-120b and a similarly sized instruct-tuned LLM, Qwen3-235B-A22B, formalized as follows:
\begin{equation}
    \label{eq4}
    \text{ratio} = \frac{1}{L} \sum \nolimits_{l=1}^L \mathbbm{1} \left( \log P_{\boldsymbol{\theta}}(\widehat y \mid \mathbf{x}, \mathbf{E}) - \log P_{\boldsymbol{\theta}}(\widehat y \mid \mathbf{x}, \mathbf{E}_{\setminus l}) < \zeta \right),
\end{equation}
where we fix $\zeta = 0$ and deem a verification step unnecessary only if it contributes negatively. As shown in Fig.~\ref{unnecessary_ratio}, when fine-tuning the same LLM, the ratio of unnecessary rationales from gpt-oss-120b is substantially higher, indeed, nearly half of all generated rationales consist entirely of useless steps (ratio $\approx 1.0$). This finding strongly suggests that \textbf{the over-verification behavior of LLMs leads to an excess of unnecessary rationales}. 

In summary, we observe that rationales produced by state-of-the-art LLMs are consistently unnecessary and insufficient. This insight motivates us to design a data filtering metric that selects only those rationales that are both necessary and sufficient for training LLMs.

%% file: S_Method.tex
\vspace{-3pt}
\section{The Proposed \baby Method}

In this section, we provide an overview of \baby in Sec.~\ref{sec3.1}, and the detailed descriptions of its two key scores in Secs.~\ref{sec3.2} and \ref{sec3.3}.

\vspace{-3pt}
\subsection{Overview of \baby} \label{sec3.1}

\begin{figure*}[t]
  \centering
  \includegraphics[width=0.935\textwidth]{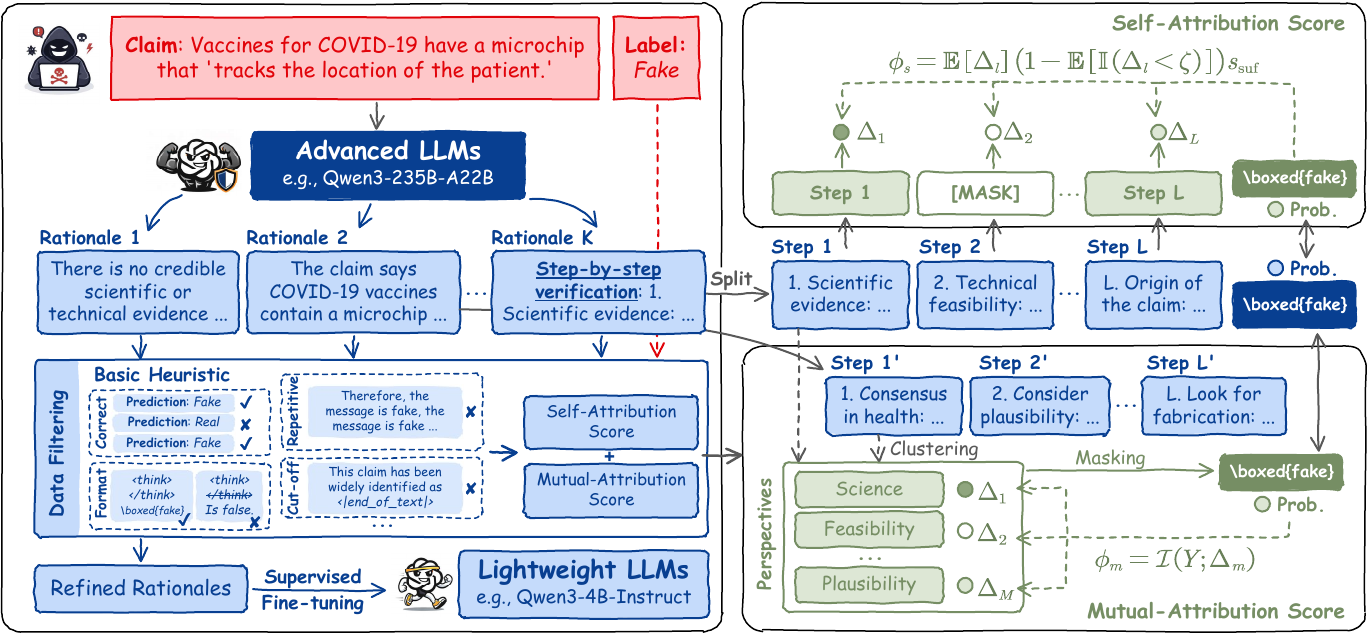}
  \vspace{-4pt}
  \caption{Overview of \baby. Given a claim, we generate $K$ rationales using advanced LLMs, which are then filtered by basic heuristics and \textit{self-attribution} and \textit{mutual-attribution} scores. 
  Finally, we use the filtered rationales to tune a lightweight LLM.}
  \label{framework}
  \vspace{-4pt}
\end{figure*}

The primary idea of \baby is to design a metric to evaluate the necessity and sufficiency of each rationale and then select higher-quality training data based on this metric. Generally, \baby comprises two main scoring components: the \textbf{self-attribution score}, which directly assesses the contribution of each verification step within a rationale to measure its necessity and sufficiency, and the \textbf{mutual-attribution score}, which generates multiple rationales for each claim and evaluates their necessity by comparing their consistency. The overview of \baby is depicted in Fig.~\ref{framework}.

\noindent
\textbf{Basic heuristic filtering.}
Formally, given the generated rationales $\left\{(\mathbf{x}_i, \mathbf{E}_{ik}, \widehat y_{ik})\right\}_{i,k=1}^{|\mathcal{D}|,K}$ formulated in Sec.~\ref{sec2}, we first apply several basic heuristic data filtering criteria. Specifically, we \textit{discard} instances where: (1) the predicted answer is incorrect, \ie $\widehat y_{ik} \neq y_i$; (2) the LLM output $\widehat y_{ik}$ does not explicitly generate either \texttt{\textbackslash boxed\{real\}} or \texttt{\textbackslash boxed\{fake\}}; (3) the rationale exceeds the token limit of 4,096 and is thus cut-off; (4) the rationale contains unrecognizable characters or unusual special symbols; or (5) the rationale exhibits repetitive or degenerate patterns, \eg copycat behavior. We denote the dataset after this initial filtering as $\mathcal{D}_0 = \left\{(\mathbf{x}_i, \mathbf{E}_{i}, \widehat y_{i})\right\}_{i=1}^{|\mathcal{D}_0|}$.

\noindent
\textbf{Self-attribution score.}
Given one instance $(\mathbf{x}_i, \mathbf{E}_{i}, \widehat y_{i}) \in \mathcal{D}_0$, we aim to calculate a self-attribution score $\phi_\text{s}$ by examining the contribution of each verification step within its rationale $\mathbf{E}_{i}$.  
Specifically, following a similar approach to that in Sec.~\ref{sec2}, we compute the necessity score $s_\text{nec}$ for each rationale by taking the average $\Delta$ as defined in Eq.~\eqref{eq:attribution} across all its verification steps, and we incorporate the ratio of unnecessary steps in Eq.~\eqref{eq4} as a penalty term. The sufficiency score $s_\text{suf}$ for each rationale is measured by the probability of its top-$\kappa$ steps in Eq.~\eqref{eq3}. The final metric $\phi_\text{s} = s_\text{nec} \times (1 - s_\text{suf})$ combines both the necessity and sufficiency scores.

\noindent
\textbf{Mutual-attribution score.}
Given a claim and its corresponding $K$ rationales $\{(\mathbf{x}_i, \mathbf{E}_{ik}, \widehat y_{ik})\}_{k=1}^K$ generated by multiple LLMs, we compute the mutual-attribution score $\phi_\text{m}$ by comparing these rationales.
Specifically, this metric follows a basic assumption: \textit{if a high-contribution verification perspective frequently appears across different rationales for the same claim, it is more likely to be necessary}. Accordingly, we cluster the verification steps within the rationales $\{\mathbf{E}_{ik}\}_{k=1}^K$  into $M$ different verification perspectives $\{\mathbf{v}_m\}_{m=1}^M$, \eg source verification and scientific validity assessment, and then compute the mutual information between each perspective and the predicted label to evaluate its contribution. A rationale that incorporates a greater number of high-contribution perspectives is deemed more necessary and is therefore assigned a higher $\phi_\text{m}$ score.

Finally, we filter samples by $(\phi_\text{s} + \phi_\text{m})/2$ to produce the final dataset $\mathcal{\widehat D} = \{(\mathbf{x}_i, \mathbf{E}_i, \widehat y_i)\}_{i=1}^{|\mathcal{\widehat D}|}$, and fine-tune LLMs by Eq.~\eqref{eq1}.
In the following sections, we describe the details of self-attribution and mutual-attribution scores, respectively.

\vspace{-3pt}
\subsection{Self-Attribution Score} \label{sec3.2}

The goal of the self-attribution score is to quantify whether a rationale provides a verification process that is both necessary and sufficient for supporting its prediction.
Given an instance $(\mathbf{x}, \mathbf{E}, \widehat y)$, where $\mathbf{x}$ is the input claim, $\widehat y$ is the predicted veracity label, and $\mathbf{E} = \{\mathbf{e}_{il}\}_{l=1}^{|\mathbf{E}|}$ denotes the rationale containing the sequence of verification steps, we aim to compute a self-attribution score $\phi_{\text{s}}(\mathbf{E})$ by evaluating the contribution of each verification step within $\mathbf{E}$.

Following the counterfactual attribution formulation in Sec.~\ref{sec2}, we measure the contribution of each verification step by removing it from the rationale and observing the change $\Delta_{l}$ in prediction confidence as in Eq.~\eqref{eq:attribution}.
Intuitively, a larger $\Delta_{l}$ indicates that the verification step is more critical to the prediction.

\noindent
\textbf{Necessity score.}
We first evaluate whether a rationale is necessary by assessing whether it contains relatively few low-contribution verification steps. The necessity score is initially defined as the average contribution across all verification steps as
\begin{equation}
s_{\text{nec}}(\mathbf{E}) = \max \Big( 0, \frac{1}{|\mathbf{E}|} \sum \nolimits_{l=1}^{|\mathbf{E}|} \Delta_{i,l} \Big).
\label{eq:self_nec}
\end{equation}

However, a rationale may still contain redundant or detrimental steps. To penalize such cases, we introduce an unnecessary-step ratio, which measures the proportion of steps whose contribution is below a threshold $\zeta=0$. The final necessity score is adjusted as
\begin{equation}
s_{\text{nec}}(\mathbf{E}) \leftarrow s_{\text{nec}}(\mathbf{E}) \cdot \Big( 1 - \frac{1}{|\mathbf{E}|} \sum \nolimits_{l=1}^{|\mathbf{E}|} \mathbbm{1} \left( \Delta_{l} < \zeta \right) \Big).
\end{equation}

\noindent
\textbf{Sufficiency score.}
Besides necessity, a high-quality rationale should contain a compact subset of steps that alone can support the prediction. We therefore identify the top-$\kappa$ steps ranked by $\Delta_{i,l}$, denoted as $\mathbf{E}_{\text{top-}\kappa}$, where $\kappa$ is empirically fixed to $3$.
The sufficiency score is defined as the prediction confidence using only these steps:
\begin{equation}
s_{\text{suf}}(\mathbf{E}) = \log P_{\boldsymbol{\theta}} \left( \widehat y \mid \mathbf{x}, \mathbf{E}_{\text{top-}\kappa} \right)
- \log P_{\boldsymbol{\theta}}(\widehat{y} \mid \mathbf{x}, \mathbf{E}).
\label{eq:self_suf}
\end{equation}
We aim to identify a \textit{minimal sufficient rationale}, \ie a higher $s_{\text{suf}}$ indicates that a small subset of steps is already sufficient for the prediction, suggesting that the remaining steps are potentially redundant.
Finally, we combine necessity and sufficiency to quantify the overall quality of the rationale:
\begin{equation}
\phi_{\text{s}}(\mathbf{E}) = s_{\text{nec}}(\mathbf{E}) \cdot \big(1 - s_{\text{suf}}(\mathbf{E})\big).
\label{eq:self_phi}
\end{equation}
This formulation encourages rationales whose verification steps are consistently important while discouraging those containing redundant or easily compressible verification processes.

\subsection{Mutual-Attribution Score} \label{sec3.3}

Given a claim $\mathbf{x}$, we assume that LLMs generate $K$ rationales and corresponding veracity predictions $\{\mathbf{E}_k, \widehat y_k\}_{k=1}^K$. Each rationale is represented as a sequence of verification steps $\mathbf{E}_k = \{\mathbf{e}_{kl}\}_{l=1}^{|\mathbf{E}_k|}$.
For each step $\mathbf{e}_{kl}$, where $k \in \{1,\ldots,K\}$ and $l \in \{1,\ldots,|\mathbf{E}_k|\}$, we obtain its semantic representation $\mathbf{h}_{kl}$ using a lightweight encoder, \eg BERT. We then use $K$-means clustering to cluster all step representations into $M$ \textbf{verification perspectives} $\mathcal{V} = \{\mathbf{v}_m\}_{m=1}^{M}$, for example, evaluating the reliability of the information source or consistency with commonsense knowledge.
Each rationale can thus be denoted as a subset of perspectives, \ie $\mathbf{E}_k \subseteq \mathcal{V}$.
Based on these perspectives, the mutual-attribution score aims to assess the contribution of each verification perspective and considers rationales that contain a greater number of high-contribution perspectives to be more necessary.

To quantify the contribution of a verification perspective $\mathbf{v}_m$ to the prediction $\widehat y$, we measure the prediction change when removing this perspective:
\begin{equation}
\Delta_{m,k} =
\log P_{\boldsymbol{\theta}} \left( \widehat y_k \mid \mathbf{x}, \mathbf{E}_k \right)
- \log P_{\boldsymbol{\theta}} \left( \widehat y_k \mid \mathbf{x}, \mathbf{E}_k \setminus \{\mathbf{v}_m\} \right).
\end{equation}

Let $\mathcal{K}_m = \{k \mid \mathbf{v}_m \in \mathbf{E}_k \}$ denote the set of rationales containing the perspective $\mathbf{v}_m$. We further define an indicator random variable
\begin{equation}
    \mathcal{A}_m =
\begin{cases}
1, & \mathbf{v}_m \in \mathbf{E}_k, \\
0, & \text{otherwise}.
\end{cases}
\end{equation}
Its empirical occurrence probability is estimated as
\begin{equation}
    P(\mathcal{A}_m=1) \approx \frac{|\mathcal{K}_m|}{K}.
\end{equation}

We define the importance of each perspective by the \textit{joint mutual information} between the prediction outcome and the perspective occurrence together with its contribution:
\begin{equation}
\phi(\mathbf{v}_m) = \mathcal{I}(Y; \mathcal{A}_m, \Delta_m).
\end{equation}
To implement it, we first use the chain rule of mutual information,
\begin{equation}
\phi(\mathbf{v}_m) = \mathcal{I}(Y; \mathcal{A}_m, \Delta_m) = \mathcal{I}(Y; \mathcal{A}_m) + \mathcal{I}(Y; \Delta_m \mid \mathcal{A}_m),
\end{equation}
where the first term is
\begin{equation}
\begin{aligned}
\mathcal{I}(Y; \mathcal{A}_m)
&= \mathbb{E}_{\mathcal{A}_m,Y} \left[ \log \frac{P(Y \mid \mathcal{A}_m)}{P(Y)} \right] \\
&\propto P(\mathcal{A}_m=1) \mathbb{E}_{Y \mid \mathcal{A}_m=1} \left[ \log P(Y \mid \mathcal{A}_m=1) \right],
\end{aligned}
\end{equation}
and the second term is
\begin{align}
\mathcal{I}(Y; \Delta_m \mid & \mathcal{A}_m) = P(\mathcal{A}_m=1) \mathbb{E}_{Y,\Delta_m \mid \mathcal{A}_m=1}
\left[ \log \frac{P(Y \mid \Delta_m,\mathcal{A}_m=1)}{P(Y \mid \mathcal{A}_m=1)} \right] \nonumber \\
& = P(\mathcal{A}_m=1) \mathbb{E}_{Y,\Delta_m \mid \mathcal{A}_m=1} [\log P(Y \mid \Delta_m,\mathcal{A}_m=1)] \nonumber \\
& \quad - P(\mathcal{A}_m=1) \mathbb{E}_{Y \mid \mathcal{A}_m=1} [\log P(Y \mid \mathcal{A}_m=1)].
\end{align}
By summing the two terms, the baseline likelihood cancels out, yielding
\begin{equation}
\phi(\mathbf{v}_m) = P(\mathcal{A}_m=1) \mathbb{E}_{Y,\Delta_m \mid \mathcal{A}_m=1} \left[ \log P(Y \mid \Delta_m,\mathcal{A}_m=1) \right].
\end{equation}
Assuming that the prediction depends on the perspective contribution only through $\Delta_m$, we approximate
\begin{equation}
    P(Y \mid \Delta_m,\mathcal{A}_m=1) \approx P(Y \mid \Delta_m).
\end{equation}
Since the true conditional distribution is unknown, we introduce a variational approximation $Q_{\boldsymbol{\Phi}}(Y \mid \Delta_m)$, which yields the evidence lower bound as 
\begin{equation}
\phi(\mathbf{v}_m) \ge P(\mathcal{A}_m=1) \mathbb{E}_{k\in \mathcal{K}_m} \left[\log Q_{\boldsymbol{\Phi}} \left( Y=\widehat y_k \mid \Delta_{m,k} \right) \right].
\end{equation}
Following a logistic stability assumption, we model
\begin{equation}
Q_{\boldsymbol{\Phi}} \left( Y=\widehat y_k  \mid  \Delta_{m,k} \right) = \left( 1 + e^{-\Delta_{m,k}} \right)^{-1}.
\end{equation}
Substituting this form and using empirical estimation, we obtain
\begin{equation}
    \begin{aligned}
    \phi(\mathbf{v}_m)
    &= -\frac{|\mathcal{K}_m|}{K} \cdot \frac{1}{|\mathcal{K}_m|} \sum \nolimits_{k\in \mathcal{K}_m} \log \left( 1 + e^{-\Delta_{m,k}} \right) \\
    &= -\frac{1}{K} \sum \nolimits_{k\in \mathcal{K}_m} \log \left( 1 + e^{-\Delta_{m,k}} \right).
    \end{aligned}
\end{equation}

Finally, the importance of a rationale is computed by aggregating the importance of its verification perspectives:
\begin{equation}
    \phi_m(\mathbf{E}_k) = \sum \nolimits_{\mathbf{v}_m \in \mathbf{E}_k} \phi(\mathbf{v}_m).
\end{equation}

%% file: S_Experiment.tex
\section{Experimental Evaluation}

In this section, we evaluate the LLM fine-tuned using our proposed \baby pipeline on their performance in MD. 

\subsection{Experimental Settings} \label{sec:app1}

\noindent
\textbf{Training datasets.}
To train a specialized LLM for explainable MD, we collect a total of 316k claims and their corresponding veracity labels from publicly available repositories. Specifically, the sources and quantities of these claims are detailed in Table~\ref{training_dataset}. After downloading the raw claims, we perform a simple deduplication step: a claim was considered a duplicate if its first 100 tokens matched those of an existing claim in the dataset.

\begin{table}[h]
\centering
\renewcommand\arraystretch{0.95}
  \caption{Sources and quantities of the training dataset.}
  \label{training_dataset}
  \small
  \setlength{\tabcolsep}{5pt}{
  \begin{tabular}{m{2.75cm}<{\centering}m{0.75cm}<{\centering}m{2.75cm}<{\centering}m{0.75cm}<{\centering}}
    \toprule
    Source & \#Num & Source & \#Num \\
    \hline
    \href{https://huggingface.co/datasets/GonzaloA/fake_news}{\color{mydarkblue} GonzaloA/fake\_news} & 40,587 & \href{https://huggingface.co/datasets/nanyy1025/covid_fake_news}{\color{mydarkblue} nanyy1025/covid\_fake.} & 10,700 \\
    \href{https://huggingface.co/datasets/ikekobby/40-percent-cleaned-preprocessed-fake-real-news}{\color{mydarkblue} ikekobby/40-percent.} & 17,957 &
    \href{https://huggingface.co/datasets/roupenminassian/twitter-misinformation}{\color{mydarkblue} roupenminassian/twitter.} & 57,761 \\
    \href{https://huggingface.co/datasets/Intel/misinformation-guard}{\color{mydarkblue} Intel/misinformation.} & 20,215 &
    \href{https://huggingface.co/datasets/pushpdeep/fake_news_combined}{\color{mydarkblue} pushpdeep/fake\_news.} & 24,123 \\
    \href{https://huggingface.co/datasets/pushpdeep/fake_news_test}{\color{mydarkblue} pushpdeep/fake\_news.} & 20,190 &
    \href{https://huggingface.co/datasets/AlexanderHolmes0/true-fake-news}{\color{mydarkblue} AlexanderHolmes0/true.} & 26,325 \\
    \href{https://huggingface.co/datasets/Hasib18/fake-news-dataset}{\color{mydarkblue} Hasib18/fake-news.} & 63,220 &
    \href{https://huggingface.co/datasets/nixbel/fakenews_train}{\color{mydarkblue} nixbel/fakenews\_train} & 23,715 \\
    \href{https://huggingface.co/datasets/lusamaki/Fake_News_Detection_System_29.5k}{\color{mydarkblue} lusamaki/Fake\_News.} & 11,859 &
    \textbf{total} & \textbf{316,652} \\
    \bottomrule
  \end{tabular} }
    \vspace{-2pt}
\end{table}

\noindent
\textbf{Evaluation dataset.}
To prevent data contamination, where evaluation samples inadvertently appear in the training data, we evaluate LLMs using the most recent MD benchmarks. \textit{MultiDis} and \textit{EUDisinfo} are collected by \citet{modzelewski2025pcot} and contain only claims published after January 2024. \textit{GossipCop++} and \textit{PolitiFact++} are compiled by \citet{su2023fake} and include both human-written and machine-generated claims, and we use only the human-written subset. Statistics for these datasets are provided in Table~\ref{mainresult}.

\begin{table*}[t]
\centering
\renewcommand\arraystretch{0.9}
  \caption{Experimental results on three LLMs fine-tuned by \baby across four MD datasets.}
  \vspace{-3pt}
  \label{mainresult}
  \small
  \setlength{\tabcolsep}{5pt}{
  \begin{tabular}{m{2.95cm}m{0.73cm}<{\centering}m{0.73cm}<{\centering}m{0.73cm}<{\centering}m{0.73cm}<{\centering}m{0.73cm}<{\centering}m{0.73cm}<{\centering}m{0.73cm}<{\centering}m{0.73cm}<{\centering}m{0.73cm}<{\centering}m{0.73cm}<{\centering}m{0.73cm}<{\centering}m{0.73cm}<{\centering}m{0.73cm}<{\centering}}
    \toprule
    \multirow{3}{*}{\quad LLMs \& Prompt} & \multicolumn{3}{c}{\textbf{\textit{GossipCop++}}} & \multicolumn{3}{c}{\textbf{\textit{PolitiFact++}}} & \multicolumn{3}{c}{\textbf{\textit{MultiDis}}} & \multicolumn{3}{c}{\textbf{\textit{EUDisinfo}}} & \multirow{3}{*}{Avg.} \\
    & \multicolumn{3}{c}{\small (\#real: 500, \#fake: 500)} & \multicolumn{3}{c}{\small (\#real: 194, \#fake: 97)} & \multicolumn{3}{c}{\small (\#real: 369, \#fake: 130)} & \multicolumn{3}{c}{\small (\#real: 241, \#fake: 118)} & \\
    \cmidrule(r){2-4} \cmidrule(r){5-7} \cmidrule(r){8-10} \cmidrule(r){11-13}
    & Acc. & F1$_\text{fake}$ & F1$_\text{real}$ & Acc. & F1$_\text{fake}$ & F1$_\text{real}$ & Acc. & F1$_\text{fake}$ & F1$_\text{real}$ & Acc. & F1$_\text{fake}$ & F1$_\text{real}$ & \\
    \hline
    \href{https://huggingface.co/Qwen/Qwen3-4B-Instruct-2507}{\color{mydarkblue} \textbf{\small Qwen3-4B-Instruct}} & 58.90 & 55.75 & 61.62 & 78.00 & 66.66 & 83.58 & 73.04 & 43.48 & 82.30 & 68.24 & 46.72 & 77.38 & 66.31 \\
    \ + CoT {\scriptsize \color{mydarkblue} (NeurIPS'22)} & 57.60 & 66.56 & 42.07 & 71.39 & 68.19 & 74.00 & 58.91 & 55.72 & 61.68 & 63.85 & 63.42 & 64.28 & 62.31 \\
    \ + ARG {\scriptsize \color{mydarkblue} (AAAI'24)} & 55.80 & 63.77 & 43.33 & 74.57 & 70.16 & 77.84 & 70.29 & 63.00 & 75.17 & 66.08 & 63.24 & 68.51 & 65.98 \\
    \ + \textsc{GenFend} {\scriptsize \color{mydarkblue} (CIKM'24)} & 60.10 & 49.17 & 67.16 & 78.00 & 57.89 & 85.11 & 74.94 & 39.02 & 84.23 & 73.81 & 54.80 & 81.56 & 67.15 \\
    \ + DMR  {\scriptsize \color{mydarkblue} (ACL'25)} & 58.75 & 66.10 & 47.31 & 69.50 & 66.91 & 71.71 & 58.66 & 55.23 & 61.61 & 63.99 & 63.66 & 64.32 & 62.31 \\
    \ + PCoT  {\scriptsize \color{mydarkblue} (ACL'25)} & 60.05 & 48.15 & 67.50 & 80.15 & 63.04 & 86.43 & 75.61 & 64.10 & 81.53 & 74.37 & 56.60 & 81.81 & 69.94 \\
    \hline
    \rowcolor{lightgrayv} \textbf{\baby-4B} {\scriptsize \color{mydarkblue}\ (Ours)} & \textbf{67.98} & \textbf{65.26} & \textbf{70.29} & \textbf{87.11} & \textbf{78.57} & \textbf{90.79} & \textbf{92.28} & \textbf{86.47} & \textbf{94.60} & \textbf{85.03} & \textbf{78.13} & \textbf{88.62} & \textbf{82.09} \\
    \rowcolor{lightgrayv} \quad $\boldsymbol{\Delta}$ \textit{v.s.} CoT & {\small \color{mydarkblue} +10.38} & {\small \color{mydarkblue} -1.30} & {\small \color{mydarkblue} +28.22} & {\small \color{mydarkblue} +15.72} & {\small \color{mydarkblue} +10.38} & {\small \color{mydarkblue} +16.79} & {\small \color{mydarkblue} +33.37} & {\small \color{mydarkblue} +30.75} & {\small \color{mydarkblue} +32.92} & {\small \color{mydarkblue} +21.18} & {\small \color{mydarkblue} +14.71} & {\small \color{mydarkblue} +24.34} & {\small \color{mydarkblue} +19.78} \\
    
    \hline
    \specialrule{0em}{0.5pt}{0.5pt}
    \hline
    \href{https://huggingface.co/Qwen/Qwen2.5-1.5B-Instruct}{\color{mydarkblue} \textbf{\small Qwen2.5-1.5B-Instruct}} & 51.82 & 9.91 & 67.12 & 73.63 & 39.21 & 83.16 & 77.45 & 31.82 & 86.49 & 66.30 & 15.09 & 78.97 & 56.75 \\
    \ + CoT {\scriptsize \color{mydarkblue} (NeurIPS'22)} & 53.52 & 17.85 & 67.60 & 75.09 & 45.90 & 83.82 & 80.06 & 49.49 & 87.58 & 67.06 & 22.33 & 79.10 & 60.78 \\
    \ + ARG {\scriptsize \color{mydarkblue} (AAAI'24)} & 54.92 & 26.74 & 67.45 & 70.27 & 42.52 & 79.95 & 80.61 & 58.52 & 87.35 & 69.56 & 36.57 & 79.98 & 62.87 \\
    \ + \textsc{GenFend} {\scriptsize \color{mydarkblue} (CIKM'24)} & 53.77 & 65.01 & 31.90 & 43.38 & 52.42 & 30.12 & 33.67 & 43.56 & 19.56 & 43.66 & 52.93 & 29.83 & 41.65 \\
    \ + DMR  {\scriptsize \color{mydarkblue} (ACL'25)} & 52.05 & 15.21 & 66.57 & 70.96 & 31.02 & 81.61 & 77.66 & 35.17 & 86.50 & 66.43 & 15.43 & 79.06 & 56.47 \\
    \ + PCoT  {\scriptsize \color{mydarkblue} (ACL'25)} & 60.50 & 63.99 & 56.26 & 61.94 & 57.92 & 65.25 & 58.82 & 53.61 & 62.97 & 61.42 & 60.32 & 62.47 & 60.46 \\
    \hline
    \rowcolor{lightgrayv} \textbf{\baby-1.5B} {\scriptsize \color{mydarkblue}\ (Ours)} & \textbf{68.50} & \textbf{69.01} & \textbf{67.97} & \textbf{84.97} & \textbf{75.59} & \textbf{89.14} & \textbf{90.63} & \textbf{83.17} & \textbf{93.51} & \textbf{83.15} & \textbf{72.31} & \textbf{87.89} & \textbf{80.49} \\
    \rowcolor{lightgrayv} \quad $\boldsymbol{\Delta}$ \textit{v.s.} CoT & {\small \color{mydarkblue} +14.98} & {\small \color{mydarkblue} +51.16} & {\small \color{mydarkblue} +0.37} & {\small \color{mydarkblue} +9.88} & {\small \color{mydarkblue} +29.69} & {\small \color{mydarkblue} +5.32} & {\small \color{mydarkblue} +10.57} & {\small \color{mydarkblue} +33.68} & {\small \color{mydarkblue} +5.93} & {\small \color{mydarkblue} +16.09} & {\small \color{mydarkblue} +49.98} & {\small \color{mydarkblue} +8.79} & {\small \color{mydarkblue} +19.71} \\
    
    \hline
    \specialrule{0em}{0.5pt}{0.5pt}
    \hline
    \href{https://huggingface.co/google/gemma-2-2b-it}{\color{mydarkblue} \textbf{\small Gemma2-2B-it}} & 54.35 & 49.59 & 58.29 & 48.36 & 37.06 & 56.22 & 58.06 & 42.47 & 67.00 & 52.86 & 36.07 & 62.66 & 51.92 \\
    \ + CoT {\scriptsize \color{mydarkblue} (NeurIPS'22)} & 56.08 & 53.13 & 58.67 & 44.02 & 41.40 & 46.42 & 44.24 & 40.32 & 47.67 & 50.00 & 40.37 & 56.95 & 48.27 \\
    \ + ARG {\scriptsize \color{mydarkblue} (AAAI'24)} & 53.26 & 53.61 & 52.90 & 55.28 & 49.58 & 59.83 & 61.77 & 49.06 & 69.40 & 52.86 & 35.28 & 62.93 & 54.65 \\
    \ + \textsc{GenFend} {\scriptsize \color{mydarkblue} (CIKM'24)} & 56.73 & 67.75 & 34.27 & 36.28 & 49.55 & 13.53 & 31.01 & 42.79 & 13.12 & 40.04 & 51.38 & 21.79 & 38.19 \\
    \ + DMR  {\scriptsize \color{mydarkblue} (ACL'25)} & 54.47 & 62.48 & 42.14 & 43.21 & 47.74 & 37.81 & 46.19 & 42.25 & 49.62 & 46.51 & 47.96 & 44.98 & 47.11 \\
    \ + PCoT  {\scriptsize \color{mydarkblue} (ACL'25)} & 57.27 & 65.60 & 43.61 & 46.45 & 53.82 & 36.28 & 44.26 & 45.03 & 43.46 & 47.55 & 50.78 & 43.88 & 48.17 \\
    \hline
    \rowcolor{lightgrayv} \textbf{\baby-2B} {\scriptsize \color{mydarkblue}\ (Ours)} & \textbf{67.12} & \textbf{65.09} & \textbf{68.93} & \textbf{81.53} & \textbf{66.46} & \textbf{87.26} & \textbf{89.58} & \textbf{82.16} & \textbf{92.64} & \textbf{78.34} & \textbf{70.12} & \textbf{83.01} & \textbf{77.69} \\
    \rowcolor{lightgrayv} \quad $\boldsymbol{\Delta}$ \textit{v.s.} CoT & {\small \color{mydarkblue} +11.04} & {\small \color{mydarkblue} +11.96} & {\small \color{mydarkblue} +10.26} & {\small \color{mydarkblue} +37.51} & {\small \color{mydarkblue} +25.06} & {\small \color{mydarkblue} +40.84} & {\small \color{mydarkblue} +45.34} & {\small \color{mydarkblue} +41.84} & {\small \color{mydarkblue} +44.97} & {\small \color{mydarkblue} +28.34} & {\small \color{mydarkblue} +29.75} & {\small \color{mydarkblue} +26.06} & {\small \color{mydarkblue} +29.42} \\
    \bottomrule
  \end{tabular} }
  \vspace{-2pt}
\end{table*}

\vspace{2pt} \noindent
\textbf{Baselines.}
We compare against 6 prompt methods specifically designed for MD, evaluating their performance across LLMs of varying sizes. These methods are as follows:
Vanilla directly prompts LLMs to make a veracity judgment;
CoT \citep{wei2022chain} instructs LLMs to reason step-by-step before generating its prediction;
ARG \citep{hu2024bad} uses LLMs to generate textual and commonsense descriptions related to the claim, which are then provided as auxiliary context for veracity prediction;
\textsc{GenFend} \citep{nan2024let} constructs diverse user personas and simulates user-generated comments to inform the prediction;
DMR \citep{jin2025disentangling} prompts LLMs to extract the key knowledge required to verify the claim.
PCoT \citep{modzelewski2025pcot} first asks LLMs to identify the persuasive strategy employed in the claim, then uses this analysis to guide the veracity assessment.
The exact prompts used for each method are provided in our code repository.

\vspace{2pt} \noindent
\textbf{Implementation details.}
During data synthesis, for the collected set of 316k claims, we employ three LLMs: Qwen3-32B, Qwen3-Next-80B-A3B, and Qwen3-235B-A22B, as rationale generators. These LLMs are selected for their favorable balance between strong performance and efficient reasoning speed (see Table~\ref{allresults}). Each LLM generates one distinct rationale per claim using a temperature of 0.6. Both data generation and model evaluation are conducted using \href{https://github.com/sgl-project/sglang}{\color{mydarkblue} SGLang} for model deployment and inference, with the maximum output length capped at 32,768 tokens.
During model fine-tuning, we performe full-parameter fine-tuning using the \href{https://github.com/hiyouga/LlamaFactory}{\color{mydarkblue} LlamaFactory} framework, setting the packing length to 32,768 to accelerate training. We use a batch size of 64 and optimize the LLM for 6 epochs with the Adam optimizer at a learning rate of $5 \times 10^{-5}$. We also experimented with training for 15 epochs, which yielded performance comparable to that of the 6-epoch setting.
During evaluation, we prompt LLMs to generate the tokens \texttt{\textbackslash boxed\{real\}} or \texttt{\textbackslash boxed\{fake\}}, and directly match these outputs against the ground-truth labels to compute accuracy and F1 score.
All LLM deployment and training were conducted on 8 NVIDIA A100 GPUs.

\subsection{Main Results}

To evaluate the effectiveness of the data generated by our \baby pipeline in improving foundation LLM performance, we train three lightweight LLMs, \eg Qwen3-4B-Instruct, Qwen2.5-1.5B-Instruct, and Gemma2-2B-it, on our curated dataset. We then evaluate their performance under carefully designed prompting strategies, including CoT \citep{wei2022chain}, ARG \citep{hu2024bad}, \textsc{GenFend} \citep{nan2024let}, DMR \citep{jin2025disentangling}, and PCoT \citep{modzelewski2025pcot}. The results across four prevalent MD benchmark datasets are reported in Table~\ref{mainresult}.
Generally, \baby consistently outperforms all baselines. As shown in Table~\ref{mainresult}, it achieves the highest accuracy and F1 scores across all three foundation LLMs. Specifically, \baby-4B achieves an average improvement of approximately 19.78\% over the CoT baseline. Moreover, the data generated by \baby achieves consistently strong performance across LLMs of three different scales, demonstrating the robustness of our data in effectively instilling verification capabilities.

\begin{table}[t]
\centering
\renewcommand\arraystretch{0.95}
  \caption{Performance compared with SOTA LLMs.}
  \label{sota_results}
  \small
  \setlength{\tabcolsep}{5pt}{
  \begin{tabular}{m{2.0cm}<{\centering}m{1.2cm}<{\centering}m{1.2cm}<{\centering}m{1.0cm}<{\centering}m{1.0cm}<{\centering}}
    \toprule
    LLM & \textit{GossipCop} & \textit{PolitiFact} & \textit{MultiDis} & \textit{EUDisinfo} \\
    \hline
    \href{https://huggingface.co/Qwen/Qwen3-32B}{\color{mydarkblue} \textbf{ Qwen3-32B}} & 62.53 & 86.96 & 86.57 & 76.11 \\
    \href{https://huggingface.co/meta-llama/Llama-3.1-70B-Instruct}{\color{mydarkblue} \textbf{ Llama-3.1-70B}} & 63.70 & 82.47 & 85.22 & 79.32 \\
    \href{https://huggingface.co/openai/gpt-oss-120b}{\color{mydarkblue}  \textbf{gpt-oss-120b}} & 58.10 & 57.13 & 65.68 & 64.83 \\
    \href{https://huggingface.co/deepseek-ai/DeepSeek-V3.2}{\color{mydarkblue} \textbf{ DeepSeek-V3.2}} & 65.15 & 89.15 & 89.07 & 78.87 \\
    \hline
    \rowcolor{lightgrayv} \textbf{\baby-4B} & \textbf{67.98} & \textbf{87.11} & \textbf{92.28} & \textbf{85.03} \\
    \rowcolor{lightgrayv} \textbf{\baby-2B} & \textbf{67.12} & \textbf{81.53} & \textbf{89.58} & \textbf{78.34} \\
    \rowcolor{lightgrayv} \textbf{\baby-1.5B} & \textbf{68.50} & \textbf{84.97} & \textbf{90.63} & \textbf{83.15} \\
    \bottomrule
  \end{tabular} }
  \vspace{-5pt}
\end{table}

\subsection{Compare with SOTA LLMs}

In this section, we compare our fine-tuned lightweight LLMs against significantly larger open-source LLMs, \eg Qwen3-32B, Llama-3.1-70B, gpt-oss-120b, and DeepSeek-V3.2. The results are presented in Table~\ref{sota_results}.
The results reveal a compelling finding: our fine-tuned LLMs significantly outperform Llama-3.1-70B and gpt-oss-120b, and are comparable with DeepSeek-V3.2, despite having less than 5\% of the parameters. 
Notably, the large reasoning model gpt-oss-120b performs poorly (\eg 57.13\% on \textit{PolitiFact++}). As analyzed in Sec.~\ref{sec2.3}, this is attributed to the over-verification behavior, where strong reasoning models tend to generate excessive verification steps, leading to a bias towards classifying claims as fake. \baby mitigates this by explicitly filtering for necessary and sufficient rationales, preventing the model from learning these redundant, noise-inducing reasoning patterns.

\subsection{Training Across Different Datasets}

We also investigate the impact of different data synthesis sources and filtering strategies. We compare \baby against three variants: (1) \textbf{w/ ALL}, using all generated rationales only filtered by correctness, (2) using rationales generated by a single LLM, filtered only by correctness, and (3) \textbf{w/ only label}, exclusively using the label format \textit{``This message is \texttt{[Label]}.''}, without rationales.

Generally, our method consistently outperforms all ablation variants. Specifically, compared to the \textit{all data} setting, which uses nearly three times more training examples, \baby achieves significantly better performance, particularly on the 4B-scale LLM. This highlights that the higher quality of data produced by \baby is more effective than simply increasing data scale. It also indicates that filtering solely based on correctness leaves substantial noise in the rationales.
In contrast to using multiple source LLMs directly, even though our data originates from these same LLMs, our pipeline, with its careful filtering strategy, yields fine-tuned LLMs that consistently surpass these source LLMs. This further underscores the importance of rigorous data curation.
Finally, LLMs trained only with the label perform poorly, demonstrating that LLMs cannot reliably infer the underlying relationship between inputs and labels. In MD, explicitly providing explainable rationales proves far more effective than relying on implicit learning from labels alone.

\begin{table}[t]
\centering
\renewcommand\arraystretch{0.95}
  \caption{Performance compared with different datasets.}
  \label{compare_training}
  \small
  \setlength{\tabcolsep}{5pt}{
  \begin{tabular}{m{1.85cm}<{\centering}m{0.45cm}<{\centering}m{1.05cm}<{\centering}m{1.05cm}<{\centering}m{0.95cm}<{\centering}m{0.95cm}<{\centering}}
    \toprule
    LLM & $N$ & \textit{GossipCop} & \textit{PolitiFact} & \textit{MultiDis} & \textit{EUDisinfo} \\
    \hline
    \rowcolor{lightgrayv} \textbf{\baby-4B} & 200k & \textbf{67.98} & \textbf{87.11} & \textbf{92.28} & \textbf{85.03} \\
    w/ Qwen3-32B & 202k & 64.75 & 83.42 & 87.17 & 79.87 \\
    w/ Qwen3-235B & 200k & 64.58 & 79.30 & 81.31 & 78.62 \\
    w/ Qwen3-Next & 197k & 65.55 & 80.58 & 88.18 & 77.86 \\
    w/ gpt-oss-120b & 180k & 58.38 & 56.79 & 78.26 & 71.31 \\
    w/ ALL & 598k & 63.85 & 77.58 & 84.12 & 77.72 \\
    w/ only label & 316k & 50.45 & 40.46 & 77.25 & 65.11 \\
    \hline
    \rowcolor{lightgrayv} \textbf{\baby-1.5B} & 200k & \textbf{68.50} & \textbf{84.97} & \textbf{90.63} & \textbf{83.15} \\
    w/ Qwen3-32B & 202k & 63.95 & 84.71 & 86.12 & 77.30 \\
    w/ Qwen3-235B & 200k & 62.18 & 74.83 & 78.56 & 77.72 \\
    w/ Qwen3-Next & 197k & 67.88 & 81.87 & 88.68 & 77.79 \\
    w/ gpt-oss-120b & 180k & 57.80 & 67.96 & 69.59 & 60.52 \\
    w/ ALL & 598k & 66.45 & 77.57 & 85.12 & 80.50 \\
    w/ only label & 316k & 49.55 & 35.99 & 62.27 & 60.86 \\
    \bottomrule
  \end{tabular} }
\end{table}

\subsection{Token Consumption}

To evaluate the efficiency of \baby, we explicitly compare the average token consumption per query during inference in Table~\ref{token_consume}.
While it consumes more tokens than vanilla prompting due to the generation of rationales, it is significantly more efficient than complex CoT or knowledge-based methods like DMR, which consumes \~2596 tokens. By internalizing the verification steps, \baby avoids the high computational overhead of multi-turn queries and reading extensive external information, making it suitable for real-time MD applications.

\subsection{Evaluation of Generated Rationales}

Finally, we assess the quality of the rationales generated by \baby. 
We follow \citet{wang2024explainable} in employing automated metrics using Qwen3-235B-A22B to evaluate the LLM-generated rationales, including \textit{Misleadingness} (M), \textit{Informativeness} (I), and \textit{Readability} (R). Misleadingness measures the degree to which the generated rationale aligns with the ground-truth veracity label, scored on a 1 (not misleading) – 5 (very misleading) scale. Informativeness assesses whether the rationale provides new knowledge beyond the claim itself, also on a 1 (not informative) – 5 (very informative) scale. Readability evaluates the fluency and readability of the rationale, again using a 1 (poor) – 5 (excellent) scale.
As shown in Table~\ref{human_evaluation}, \baby-4B achieves rationale quality scores that are competitive with, and occasionally superior to, the teacher LLMs, \eg Qwen3-32B. This indicates that our fine-tuning process does not merely memorize labels but learns to construct coherent, logically sound verification chains.
Additionally, some representative cases are shown in Table~\ref{case}.

\begin{table}[t]
\centering
\renewcommand\arraystretch{0.95}
  \caption{Token consumption of different methods.}
  \vspace{-3pt}
  \label{token_consume}
  \small
  \setlength{\tabcolsep}{5pt}{
  \begin{tabular}{m{1.05cm}<{\centering}m{1.25cm}<{\centering}m{1.45cm}<{\centering}m{1.45cm}<{\centering}m{1.4cm}<{\centering}}
    \toprule
    Method & Qwen3-4B & Gemma2-2B & Qwen3-1.5B & Llama3-70B \\
    \hline
    Vanilla & 267.52 & 223.28 & 85.24 & 191.29 \\
    CoT & 1154.82 & 712.63 & 132.22 & 937.85 \\
    ARG & 678.81 & 617.80 & 342.12 & 780.74 \\
    \textsc{GenFend} & 663.08 & 982.52 & 3027.22 & 414.54 \\
    DMR & 2596.75 & 1203.38 & 1100.31 & 1387.56 \\
    PCoT & 1369.48 & 1023.23 & 978.12 & 1146.28 \\
    \rowcolor{lightgrayv} \textbf{\baby} & 676.65 & 502.64 & 491.20 & - \\
    \bottomrule
  \end{tabular} }
\end{table}

\begin{table}[t]
\centering
\renewcommand\arraystretch{0.95}
  \caption{Evaluation of rationales generated by different LLMs.}
  \vspace{-3pt}
  \label{human_evaluation}
  \small
  \setlength{\tabcolsep}{5pt}{
  \begin{tabular}{m{1.9cm}<{\centering}m{0.6cm}<{\centering}m{0.6cm}<{\centering}m{0.6cm}<{\centering}m{0.6cm}<{\centering}m{0.6cm}<{\centering}m{0.6cm}<{\centering}}
    \toprule
    \multirow{2}{*}{LLM} & \multicolumn{3}{c}{\textit{GossipCop++}} & \multicolumn{3}{c}{\textit{EUDisinfo}} \\
    \cmidrule(r){2-4} \cmidrule(r){5-7}
     & M & I & R & M & I & R \\
    \hline
    \rowcolor{lightgrayv} \textbf{\baby-1.5B} & 2.17 & 3.80 & 4.13 & 1.95 & 4.09 & 4.17 \\
    \rowcolor{lightgrayv} \textbf{\baby-2B} & 2.24 & 3.75 & 4.14 & 2.02 & 4.01 & 4.13 \\
    \rowcolor{lightgrayv} \textbf{\baby-4B} & 2.13 & 3.69 & 4.21 & 1.98 & 4.07 & 4.20 \\
    \hline
    Qwen3-32B & 2.30 & 3.67 & 4.05 & 2.11 & 3.97 & 4.07 \\
    Llama3-70B & 2.28 & 3.70 & 4.18 & 1.98 & 4.02 & 4.11 \\
    gpt-oss-120b & 2.61 & 3.52 & 3.88 & 2.32 & 3.92 & 4.00 \\
    DeepSeek-V3.2 & 2.02 & 3.90 & 4.27 & 2.01 & 4.20 & 4.31 \\
    \bottomrule
  \end{tabular} }
\end{table}

\begin{table*}[h]
\centering
\renewcommand\arraystretch{1.0}
  \caption{A representative case from the \textit{MultiDis} dataset generated by different LLMs.}
  \vspace{-3pt}
  \label{case}
  \footnotesize
  \setlength{\tabcolsep}{5pt}{
  \begin{tabular}{p{\textwidth}}
    \toprule
    \textbf{Article}: This wouldn't be happening without climate change', said {\color{mydarkblue}the head of the United States' Coral Reef Watch Program}
Ocean temperatures that have gone haywire could make the current global coral bleaching the worst in history.
...More than {\color{mydarkblue}62.9\%} of the world's coral reefs are badly hurting from a bleaching event that began last year... {\color{mydarkblue}nearing the record of 65.7\% in 2017}...
...{\color{mydarkblue}99.7\% of the coral reefs} have been hit with "very very severe" losses in staghorn and elkhorn species...
...{\color{mydarkblue}Thailand shutting off a tourist-laden island} to try to save the coral there.
Meteorologists say a {\color{mydarkblue}La Nina}... forecast to develop soon ..., but Manzello said it may be {\color{mydarkblue}too little and too late}.
“I still am very worried...” Manzello said.
“This wouldn't be happening without climate change...” Manzello said.
On top of that are changes in {\color{mydarkblue}El Nino}...; {\color{mydarkblue}reduced sulfur pollution from ships} and {\color{mydarkblue}an undersea volcano eruption}.
Former top NASA climate scientist {\color{mydarkblue}James Hansen} said “acceleration of global warming is now hard to deny”...
...scientists are hoping for {\color{mydarkblue}a few hurricanes} as they cool the oceans...
...{\color{mydarkblue}the world's oceans last month broke a record} for the hottest April on record...
Coral reefs are key to seafood production... {\color{mydarkblue}loss of coral is one of the big tipping points}... {\color{mydarkblue}countries agreed to try to hold to} in the 2015 Paris climate agreement. \\
    \hline
    \textbf{Llama3.1-70B} (534 tokens): The message appears to be an article from a reputable source, citing {\color{mydarkblue}NOAA} and {\color{mydarkblue}Climate Central}.
The information is specific, with {\color{mydarkblue}quotes from experts} adding credibility.
Mentions recent records and trends like {\color{mydarkblue}the hottest April on record}, suggesting the info is up-to-date.
...
No obvious contradictions found, therefore: \color{mydarkblue}{real} \\
    \hline
    \textbf{gpt-oss-120b} (1048 tokens): We need to determine if the message is likely real or fake. The content is a news-like article quoting NOAA coral reef watch program head Derek Manzello...
Check for inconsistencies: {\color{mydarkblue}"head of the United States' Coral Reef Watch Program"} — NOAA's Coral Reef Watch is a program, but the head is likely a scientist; Derek Manzello is a NOAA coral reef watch coordinator.
The quotes seem plausible. However, some statements: {\color{mydarkblue}"99.7\% of the coral reefs have been hit..."} seems extreme; also {\color{mydarkblue}"Thailand shutting off a tourist-laden island"}... plausible but needs checking.
...Also {\color{mydarkblue}"reduced sulfur pollution from ships"} and {\color{mydarkblue}"an undersea volcano eruption"} — sentence is odd mixing topics.
...The phrase {\color{mydarkblue}"very very severe"} seems informal. Also {\color{mydarkblue}"the reverse of La Nina"} is simplistic.
Thus likely {\color{mydarkblue}fake}. \\
    \hline
    \textbf{Qwen3-4B-Think-2507} (5,363 tokens): Okay, step by step...
Claims: {\color{mydarkblue}62.9\% of world's coral reefs} badly hurt (nearing {\color{mydarkblue}65.7\% in 2017}); {\color{mydarkblue}99.7\% of Atlantic reefs} "very very severe" losses; {\color{mydarkblue}Thailand shutting off island}; {\color{mydarkblue}La Niña forecast too little, too late}; {\color{mydarkblue}hurricanes to cool reefs}; {\color{mydarkblue}hottest April on record}.
{\color{mydarkblue}NOAA} has Coral Reef Watch program; data shows major bleaching. 62.9\% figure plausible.
{\color{mydarkblue}99.7\% figure catches my eye} — extremely high. NOAA data shows Florida severe damage but {\color{mydarkblue}not 99.7\%}.
...{\color{mydarkblue}62 countries seeing damaged coral} maybe high.
Scientists hope for hurricanes — plausible.
Thailand closure real.
Key issue: {\color{mydarkblue}99.7\% likely fabricated to alarm people}. NOAA uses "over 50\%" or "severe in many areas" not exact 99.7\%.
Given overall alignment but {\color{mydarkblue}99.7\% unsupported}, leaning {\color{mydarkblue}fake}. \\

\hline
\textbf{\baby-4B} (678 tokens): Let’s analyze this message step by step for veracity, using known scientific facts, credible sources, and plausibility.
1. Source Attribution: Derek Manzello, NOAA Coral Reef Watch Program
- Fact Check: Derek Manzello {\color{mydarkblue}{is a real person}} and was indeed the Coordinator of {\color{mydarkblue}{NOAA’s Coral Reef Watch Program}} during the relevant time period (2016–2018). He frequently appeared in media briefings and scientific reports on coral bleaching.
- Verdict: Real.
2. “This wouldn't be happening without climate change”
- This is a direct quote that matches {\color{mydarkblue}{Manzello’s public statements during the 2016–2017 mass coral bleaching events}}.
- {\color{mydarkblue}{NOAA}}, {\color{mydarkblue}{IPCC}}, and other scientific bodies have consistently stated that mass coral bleaching events are driven by ocean warming due to {\color{mydarkblue}{anthropogenic climate change}}.
- Verdict: Accurate.
...
Conclusion:
Every claim in the message:
- Is attributed to {\color{mydarkblue}{real people}} ({\color{mydarkblue}{Manzello}}, {\color{mydarkblue}{Hansen}}, {\color{mydarkblue}{McNoldy}}, {\color{mydarkblue}{Klotzbach}}),
- Matches {\color{mydarkblue}{verified scientific reports}} from {\color{mydarkblue}{NOAA}}, {\color{mydarkblue}{IPCC}}, and peer-reviewed literature,
- Uses {\color{mydarkblue}{accurate statistics from 2016–2017}},
- Reflects {\color{mydarkblue}{actual public statements}} made during that time.
There are no exaggerations, no fabrications, and no misleading edits.
Final Answer: {\color{mydarkblue}real} \\
    \bottomrule
  \end{tabular} }
    \vspace{-2pt}
\end{table*}

%% file: S_RelatedWorks.tex
\section{Related Works}

In this section, we summarize the related literature on LLM-based misinformation detection and advanced CoT reasoning techniques.

\subsection{LLM-based Misinformation Detection}

Typically, previous MD methods focus on training \textit{small discriminative models} that leverage external features, \eg linguistic style \citep{przybyla2020capturing}, propagation patterns \citep{bian2020rumor,wu2023decor}, and more modalities \citep{qi2023fakesv}. However, these models heavily depend on labeled data and possess limited world knowledge, making them hard to handle rapidly evolving, out-of-distribution news events.
Recently, partial studies have begun leveraging \textit{large generative models} to aid in misinformation detection and explanation. 
One line of research leverages LLMs to retrieve or synthesize external information that augments small discriminative models in their decision-making \citep{hu2024bad,nan2024let,wang2025collaboration}. For example, ARG \citep{hu2024bad} generates textual descriptions and commonsense rationales, and DELL \citep{wan2024dell} simulates user comments to serve as auxiliary knowledge that aids smaller detectors.
Another line employs carefully crafted prompts or multi-agent collaboration to enable LLMs to directly produce veracity judgments and explanatory justifications \citep{hong2025following,modzelewski2025pcot,tong2025generate}. For example, TED \citep{liu2025the} orchestrates multiple agents in a structured debate process, and DYNAMO \citep{jin2025a} decomposes complex articles into easily verifiable sub-questions through Monte Carlo tree search, thereby enabling deep logical reasoning.

\subsection{Long Chain-of-Thought Reasoning}

Recently, the introduction of LLMs has shifted research focus toward enhancing long chain-of-thought reasoning \citep{yuan2024instance,guo2025deepseek,yang2026toward}. This paradigm involves generating reasoning trajectories prior to producing a final answer \citep{wei2022chain,kojima2022large}, which closely mirrors explainable MD. 
Existing pipelines typically leverage supervised fine-tuning and reinforcement learning techniques. In reinforcement learning-based frameworks, meticulously designed rewards, \eg those utilizing label correctness as a signal \citep{shao2024deepseekmath}, are employed to incentivize LLMs to generate reasoning trajectories. However, in explainable MD, where ground-truth labels are binary, such rewards are inherently sparse. This often results in training instability and may even lead to degraded performance \citep{du2023guiding}.

To address these limitations, our work explores an alternative direction that aggregates reasoning traces generated by diverse models and develops a filtering mechanism to extract high-quality trajectories for the supervised fine-tuning of LLMs \citep{ho2023large,wang2023scott,kou2025rethinking,yuan2026differential}.
For example, existing approaches often employ heuristic rules, \eg response format \citep{muennighoff2025s1,huang2025loong}, question difficulty \citep{ye2025limo}, or answer accuracy \citep{lei2025learning,chen2025skip}, or assess LLMs' adaptability to the responses \citep{zhang2025the,liu2025where,yan2026distribution}. In contrast, our work prioritizes data selection through counterfactual attribution by selectively masking intermediate reasoning steps to observe their impact on the final output \citep{chen2025skip}.

%% file: S_Conclusion.tex
\section{Conclusion}

In this work, we address the growing need for transparency in MD by shifting the focus from black-box classification to explainable MD. While the prevailing trend leverages LLMs through prompting or naive fine-tuning, our experiments reveal that existing data synthesis pipelines suffer from two critical bottlenecks: \textbf{insufficient} rationales caused by coarse-grained labels and \textbf{unnecessary} rationales caused by the over-verification behavior. These issues introduce noise and verbosity that hinder the training of LLMs.
To overcome these issues, we introduce \textbf{\baby}, a novel data-centric pipeline. By implementing a perplexity-based metric grounded in self-attribution and mutual-attribution, \baby effectively identifies and selects rationales that are both necessary and sufficient. Our experimental results demonstrate that fine-tuning lightweight LLMs (\eg 1.5B to 4B parameters) using our filtered dataset leads to a significant performance boost, approximately 22.97\% over foundation models, achieving results comparable to much larger state-of-the-art open-source LLMs.
Ultimately, by releasing our raw dataset and the refined \baby version, we provide a robust resource for the community to develop more transparent MD systems.

%% file: S_Appendix.tex
\appendix

\begin{table*}
\centering
\renewcommand\arraystretch{0.942}
  \caption{Experimental results on more advanced LLMs across four MD benchmarks.}
  \label{allresults}
  \small
  \setlength{\tabcolsep}{5pt}{
  \begin{tabular}{m{2.70cm}m{0.95cm}<{\centering}m{0.65cm}<{\centering}m{0.65cm}<{\centering}m{0.65cm}<{\centering}m{0.65cm}<{\centering}m{0.65cm}<{\centering}m{0.65cm}<{\centering}m{0.65cm}<{\centering}m{0.65cm}<{\centering}m{0.65cm}<{\centering}m{0.65cm}<{\centering}m{0.65cm}<{\centering}m{0.65cm}<{\centering}m{0.65cm}<{\centering}}
    \toprule
    \multirow{2}{*}{\quad LLMs \& Prompt} & \multirow{2}{*}{\#param} & \multicolumn{3}{c}{\textit{GossipCop++}} & \multicolumn{3}{c}{\textit{PolitiFact++}} & \multicolumn{3}{c}{\textit{MultiDis}} & \multicolumn{3}{c}{\textit{EUDisinfo}} & \multirow{2}{*}{Avg.} \\
    \cmidrule(r){3-5} \cmidrule(r){6-8} \cmidrule(r){9-11} \cmidrule(r){12-14}
    & & Acc. & F1$_\text{fake}$ & F1$_\text{real}$ & Acc. & F1$_\text{fake}$ & F1$_\text{real}$ & Acc. & F1$_\text{fake}$ & F1$_\text{real}$ & Acc. & F1$_\text{fake}$ & F1$_\text{real}$ & \\
    \hline
    \multicolumn{15}{c}{\textbf{$>$ 100B parameters}} \\
    \href{https://huggingface.co/deepseek-ai/DeepSeek-V3.2}{\color{mydarkblue} \textbf{\footnotesize DeepSeek-V3.2}} & 685B & \cellcolor{mydarkblue!25}68.97 & \cellcolor{mydarkblue!17}61.18 & \cellcolor{mydarkblue!25}74.16 & \cellcolor{mydarkblue!24}87.89 & \cellcolor{mydarkblue!24}79.71 & \cellcolor{mydarkblue!24}91.37 & \cellcolor{mydarkblue!24}90.03 & \cellcolor{mydarkblue!23}82.89 & \cellcolor{mydarkblue!23}92.97 & \cellcolor{mydarkblue!24}84.12 & \cellcolor{mydarkblue!22}77.34 & \cellcolor{mydarkblue!23}87.78 & \cellcolor{mydarkblue!24}81.53 \\
    \ + CoT {\scriptsize \color{mydarkblue} (NeurIPS'22)} & 685B & \cellcolor{mydarkblue!18}65.15 & \cellcolor{mydarkblue!7}50.06 & \cellcolor{mydarkblue!24}73.23 & \cellcolor{mydarkblue!25}89.15 & \cellcolor{mydarkblue!25}80.95 & \cellcolor{mydarkblue!25}92.41 & \cellcolor{mydarkblue!23}89.07 & \cellcolor{mydarkblue!21}80.28 & \cellcolor{mydarkblue!23}92.44 & \cellcolor{mydarkblue!21}78.87 & \cellcolor{mydarkblue!8}61.63 & \cellcolor{mydarkblue!21}85.42 & \cellcolor{mydarkblue!21}78.22 \\
    \ +  ARG {\scriptsize \color{mydarkblue} (AAAI'24)} & 685B & \cellcolor{mydarkblue!22}67.12 & \cellcolor{mydarkblue!12}56.15 & \cellcolor{mydarkblue!25}73.71 & \cellcolor{mydarkblue!25}88.57 & \cellcolor{mydarkblue!25}81.08 & \cellcolor{mydarkblue!25}91.82 & \cellcolor{mydarkblue!24}91.18 & \cellcolor{mydarkblue!24}84.64 & \cellcolor{mydarkblue!24}93.82 & \cellcolor{mydarkblue!24}83.22 & \cellcolor{mydarkblue!19}74.44 & \cellcolor{mydarkblue!23}87.51 & \cellcolor{mydarkblue!24}81.11 \\
    \ + \textsc{GenFend} {\scriptsize \color{mydarkblue} (CIKM'24)} & 685B & \cellcolor{mydarkblue!21}66.62 & \cellcolor{mydarkblue!10}53.24 & \cellcolor{mydarkblue!25}74.05 & \cellcolor{mydarkblue!24}87.63 & \cellcolor{mydarkblue!24}79.60 & \cellcolor{mydarkblue!24}91.12 & \cellcolor{mydarkblue!25}92.64 & \cellcolor{mydarkblue!25}86.55 & \cellcolor{mydarkblue!25}94.93 & \cellcolor{mydarkblue!25}84.68 & \cellcolor{mydarkblue!23}77.91 & \cellcolor{mydarkblue!24}88.27 & \cellcolor{mydarkblue!24}81.44 \\
    \ + DMR {\scriptsize \color{mydarkblue} (ACL'25)} & 685B & \cellcolor{mydarkblue!21}66.75 & \cellcolor{mydarkblue!12}55.70 & \cellcolor{mydarkblue!25}73.39 & \cellcolor{mydarkblue!25}88.49 & \cellcolor{mydarkblue!25}80.91 & \cellcolor{mydarkblue!25}91.76 & \cellcolor{mydarkblue!24}89.73 & \cellcolor{mydarkblue!23}82.52 & \cellcolor{mydarkblue!23}92.73 & \cellcolor{mydarkblue!23}82.73 & \cellcolor{mydarkblue!19}74.33 & \cellcolor{mydarkblue!23}86.99 & \cellcolor{mydarkblue!24}80.50 \\
    \ + PCoT {\scriptsize \color{mydarkblue} (ACL'25)} & 685B & \cellcolor{mydarkblue!19}65.80 & \cellcolor{mydarkblue!8}51.73 & \cellcolor{mydarkblue!25}73.52 & \cellcolor{mydarkblue!21}84.28 & \cellcolor{mydarkblue!19}74.41 & \cellcolor{mydarkblue!23}88.65 & \cellcolor{mydarkblue!24}91.13 & \cellcolor{mydarkblue!23}83.22 & \cellcolor{mydarkblue!24}93.97 & \cellcolor{mydarkblue!25}85.58 & \cellcolor{mydarkblue!22}76.87 & \cellcolor{mydarkblue!25}89.53 & \cellcolor{mydarkblue!23}79.89 \\
    \hline
    \href{https://huggingface.co/Qwen/Qwen3-235B-A22B-Instruct-2507}{\color{mydarkblue} \textbf{\footnotesize Qwen3-235B-A22B-Ins.}} & 235B & \cellcolor{mydarkblue!18}64.98 & \cellcolor{mydarkblue!22}66.44 & \cellcolor{mydarkblue!18}63.37 & \cellcolor{mydarkblue!22}84.88 & \cellcolor{mydarkblue!22}77.83 & \cellcolor{mydarkblue!23}88.53 & \cellcolor{mydarkblue!23}88.43 & \cellcolor{mydarkblue!22}81.02 & \cellcolor{mydarkblue!22}91.68 & \cellcolor{mydarkblue!19}75.56 & \cellcolor{mydarkblue!15}68.97 & \cellcolor{mydarkblue!16}79.84 & \cellcolor{mydarkblue!21}77.63 \\
    \ + CoT {\scriptsize \color{mydarkblue} (NeurIPS'22)} & 235B & \cellcolor{mydarkblue!15}63.15 & \cellcolor{mydarkblue!14}57.62 & \cellcolor{mydarkblue!21}67.40 & \cellcolor{mydarkblue!21}84.02 & \cellcolor{mydarkblue!17}71.99 & \cellcolor{mydarkblue!23}88.82 & \cellcolor{mydarkblue!22}86.52 & \cellcolor{mydarkblue!20}78.18 & \cellcolor{mydarkblue!21}90.25 & \cellcolor{mydarkblue!20}77.16 & \cellcolor{mydarkblue!14}67.84 & \cellcolor{mydarkblue!18}82.29 & \cellcolor{mydarkblue!20}76.27 \\
    \ +  ARG {\scriptsize \color{mydarkblue} (AAAI'24)} & 235B & \cellcolor{mydarkblue!20}66.20 & \cellcolor{mydarkblue!17}60.74 & \cellcolor{mydarkblue!23}70.32 & \cellcolor{mydarkblue!21}84.19 & \cellcolor{mydarkblue!21}76.04 & \cellcolor{mydarkblue!22}88.21 & \cellcolor{mydarkblue!24}89.63 & \cellcolor{mydarkblue!22}81.27 & \cellcolor{mydarkblue!23}92.83 & \cellcolor{mydarkblue!22}80.99 & \cellcolor{mydarkblue!17}71.71 & \cellcolor{mydarkblue!21}85.68 & \cellcolor{mydarkblue!22}78.98 \\
    \ + \textsc{GenFend} {\scriptsize \color{mydarkblue} (CIKM'24)} & 235B & \cellcolor{mydarkblue!16}64.10 & \cellcolor{mydarkblue!19}62.97 & \cellcolor{mydarkblue!19}65.16 & \cellcolor{mydarkblue!20}82.38 & \cellcolor{mydarkblue!19}73.75 & \cellcolor{mydarkblue!21}86.74 & \cellcolor{mydarkblue!24}89.98 & \cellcolor{mydarkblue!23}82.82 & \cellcolor{mydarkblue!23}92.93 & \cellcolor{mydarkblue!20}77.72 & \cellcolor{mydarkblue!15}69.35 & \cellcolor{mydarkblue!18}82.49 & \cellcolor{mydarkblue!21}77.53 \\
    \ + DMR {\scriptsize \color{mydarkblue} (ACL'25)} & 235B & \cellcolor{mydarkblue!21}66.70 & \cellcolor{mydarkblue!23}67.34 & \cellcolor{mydarkblue!20}66.04 & \cellcolor{mydarkblue!21}83.68 & \cellcolor{mydarkblue!21}76.19 & \cellcolor{mydarkblue!22}87.58 & \cellcolor{mydarkblue!21}84.82 & \cellcolor{mydarkblue!19}76.99 & \cellcolor{mydarkblue!20}88.67 & \cellcolor{mydarkblue!23}83.01 & \cellcolor{mydarkblue!22}77.66 & \cellcolor{mydarkblue!22}86.29 & \cellcolor{mydarkblue!22}78.75 \\
    \ + PCoT {\scriptsize \color{mydarkblue} (ACL'25)} & 235B & \cellcolor{mydarkblue!21}66.77 & \cellcolor{mydarkblue!15}59.04 & \cellcolor{mydarkblue!24}72.05 & \cellcolor{mydarkblue!22}85.48 & \cellcolor{mydarkblue!22}77.79 & \cellcolor{mydarkblue!23}89.22 & \cellcolor{mydarkblue!24}90.08 & \cellcolor{mydarkblue!23}82.90 & \cellcolor{mydarkblue!23}93.01 & \cellcolor{mydarkblue!25}84.96 & \cellcolor{mydarkblue!23}78.57 & \cellcolor{mydarkblue!24}88.41 & \cellcolor{mydarkblue!24}80.69 \\
    \hline
    \href{https://huggingface.co/openai/gpt-oss-120b}{\color{mydarkblue} \footnotesize \textbf{gpt-oss-120b} (high effort)} & 120B & \cellcolor{mydarkblue!6}58.05 & \cellcolor{mydarkblue!23}66.88 & \cellcolor{mydarkblue!6}42.81 & \cellcolor{mydarkblue!2}57.47 & \cellcolor{mydarkblue!4}57.22 & \cellcolor{mydarkblue!3}57.73 & \cellcolor{mydarkblue!13}68.29 & \cellcolor{mydarkblue!9}60.71 & \cellcolor{mydarkblue!8}73.41 & \cellcolor{mydarkblue!12}64.35 & \cellcolor{mydarkblue!4}56.46 & \cellcolor{mydarkblue!6}69.81 & \cellcolor{mydarkblue!5}61.10 \\
    \ + CoT {\scriptsize \color{mydarkblue} (NeurIPS'22)} & 120B & \cellcolor{mydarkblue!6}58.10 & \cellcolor{mydarkblue!23}67.18 & \cellcolor{mydarkblue!5}42.09 & \cellcolor{mydarkblue!2}57.13 & \cellcolor{mydarkblue!4}57.09 & \cellcolor{mydarkblue!3}57.17 & \cellcolor{mydarkblue!12}65.68 & \cellcolor{mydarkblue!7}58.51 & \cellcolor{mydarkblue!6}70.74 & \cellcolor{mydarkblue!12}64.83 & \cellcolor{mydarkblue!4}56.50 & \cellcolor{mydarkblue!7}70.49 & \cellcolor{mydarkblue!5}60.46 \\
    \ +  ARG {\scriptsize \color{mydarkblue} (AAAI'24)} & 120B & \cellcolor{mydarkblue!10}60.75 & \cellcolor{mydarkblue!23}67.22 & \cellcolor{mydarkblue!11}51.09 & \cellcolor{mydarkblue!4}61.17 & \cellcolor{mydarkblue!6}59.21 & \cellcolor{mydarkblue!6}62.95 & \cellcolor{mydarkblue!15}71.09 & \cellcolor{mydarkblue!10}62.06 & \cellcolor{mydarkblue!11}76.65 & \cellcolor{mydarkblue!15}68.31 & \cellcolor{mydarkblue!5}58.30 & \cellcolor{mydarkblue!10}74.45 & \cellcolor{mydarkblue!9}64.44 \\
    \ + \textsc{GenFend} {\scriptsize \color{mydarkblue} (CIKM'24)} & 120B & \cellcolor{mydarkblue!5}57.95 & \cellcolor{mydarkblue!14}57.46 & \cellcolor{mydarkblue!15}58.43 & \cellcolor{mydarkblue!2}58.42 & \cellcolor{mydarkblue!1}52.92 & \cellcolor{mydarkblue!6}62.77 & \cellcolor{mydarkblue!6}54.06 & \cellcolor{mydarkblue!0}34.27 & \cellcolor{mydarkblue!1}64.69 & \cellcolor{mydarkblue!7}55.78 & \cellcolor{mydarkblue!0}43.35 & \cellcolor{mydarkblue!0}63.74 & \cellcolor{mydarkblue!0}55.32 \\
    \ + DMR {\scriptsize \color{mydarkblue} (ACL'25)} & 120B & \cellcolor{mydarkblue!5}57.72 & \cellcolor{mydarkblue!20}64.04 & \cellcolor{mydarkblue!9}48.71 & \cellcolor{mydarkblue!5}61.34 & \cellcolor{mydarkblue!4}57.22 & \cellcolor{mydarkblue!7}64.73 & \cellcolor{mydarkblue!13}67.64 & \cellcolor{mydarkblue!6}56.05 & \cellcolor{mydarkblue!9}74.39 & \cellcolor{mydarkblue!12}64.90 & \cellcolor{mydarkblue!0}52.72 & \cellcolor{mydarkblue!8}72.09 & \cellcolor{mydarkblue!6}61.80 \\
    \ + PCoT {\scriptsize \color{mydarkblue} (ACL'25)} & 120B & \cellcolor{mydarkblue!17}64.70 & \cellcolor{mydarkblue!22}65.83 & \cellcolor{mydarkblue!18}63.50 & \cellcolor{mydarkblue!10}68.90 & \cellcolor{mydarkblue!10}64.30 & \cellcolor{mydarkblue!12}72.45 & \cellcolor{mydarkblue!21}84.77 & \cellcolor{mydarkblue!19}76.69 & \cellcolor{mydarkblue!20}88.69 & \cellcolor{mydarkblue!22}80.78 & \cellcolor{mydarkblue!19}73.61 & \cellcolor{mydarkblue!21}84.88 & \cellcolor{mydarkblue!18}74.09 \\
    \hline
    \specialrule{0em}{0.5pt}{0.5pt}
    \hline
    \multicolumn{15}{c}{\textbf{$>$ 10B, $<$ 100B parameters}} \\
    \href{https://huggingface.co/Qwen/Qwen3-Next-80B-A3B-Instruct}{\color{mydarkblue} \textbf{\footnotesize Qwen3-Next-80B-A3B}} & 80B & \cellcolor{mydarkblue!13}62.30 & \cellcolor{mydarkblue!18}62.15 & \cellcolor{mydarkblue!18}62.45 & \cellcolor{mydarkblue!22}84.97 & \cellcolor{mydarkblue!22}78.26 & \cellcolor{mydarkblue!23}88.51 & \cellcolor{mydarkblue!23}88.93 & \cellcolor{mydarkblue!22}81.96 & \cellcolor{mydarkblue!23}92.01 & \cellcolor{mydarkblue!23}82.80 & \cellcolor{mydarkblue!21}76.76 & \cellcolor{mydarkblue!22}86.35 & \cellcolor{mydarkblue!22}78.95 \\
    \ + CoT {\scriptsize \color{mydarkblue} (NeurIPS'22)} & 80B & \cellcolor{mydarkblue!20}65.92 & \cellcolor{mydarkblue!19}63.01 & \cellcolor{mydarkblue!21}68.41 & \cellcolor{mydarkblue!23}86.51 & \cellcolor{mydarkblue!24}79.64 & \cellcolor{mydarkblue!23}89.92 & \cellcolor{mydarkblue!23}89.13 & \cellcolor{mydarkblue!22}82.29 & \cellcolor{mydarkblue!23}92.16 & \cellcolor{mydarkblue!21}79.25 & \cellcolor{mydarkblue!18}73.20 & \cellcolor{mydarkblue!19}83.07 & \cellcolor{mydarkblue!22}79.38 \\
    \ +  ARG {\scriptsize \color{mydarkblue} (AAAI'24)} & 80B & \cellcolor{mydarkblue!15}63.18 & \cellcolor{mydarkblue!16}60.20 & \cellcolor{mydarkblue!20}65.74 & \cellcolor{mydarkblue!22}84.79 & \cellcolor{mydarkblue!22}77.85 & \cellcolor{mydarkblue!22}88.42 & \cellcolor{mydarkblue!22}86.77 & \cellcolor{mydarkblue!20}78.64 & \cellcolor{mydarkblue!21}90.42 & \cellcolor{mydarkblue!21}79.32 & \cellcolor{mydarkblue!17}71.14 & \cellcolor{mydarkblue!20}83.88 & \cellcolor{mydarkblue!21}77.53 \\
    \ + \textsc{GenFend} {\scriptsize \color{mydarkblue} (CIKM'24)} & 80B & \cellcolor{mydarkblue!18}64.95 & \cellcolor{mydarkblue!19}63.09 & \cellcolor{mydarkblue!20}66.63 & \cellcolor{mydarkblue!23}85.82 & \cellcolor{mydarkblue!23}78.93 & \cellcolor{mydarkblue!23}89.32 & \cellcolor{mydarkblue!23}87.88 & \cellcolor{mydarkblue!21}80.42 & \cellcolor{mydarkblue!22}91.22 & \cellcolor{mydarkblue!21}78.32 & \cellcolor{mydarkblue!7}59.70 & \cellcolor{mydarkblue!21}85.17 & \cellcolor{mydarkblue!21}77.62 \\
    \ + DMR {\scriptsize \color{mydarkblue} (ACL'25)} & 80B & \cellcolor{mydarkblue!11}61.08 & \cellcolor{mydarkblue!21}64.78 & \cellcolor{mydarkblue!14}56.50 & \cellcolor{mydarkblue!18}78.95 & \cellcolor{mydarkblue!18}73.34 & \cellcolor{mydarkblue!19}82.61 & \cellcolor{mydarkblue!14}69.54 & \cellcolor{mydarkblue!10}62.79 & \cellcolor{mydarkblue!9}74.22 & \cellcolor{mydarkblue!17}72.21 & \cellcolor{mydarkblue!14}68.46 & \cellcolor{mydarkblue!11}75.17 & \cellcolor{mydarkblue!14}69.97 \\
    \ + PCoT {\scriptsize \color{mydarkblue} (ACL'25)} & 80B & \cellcolor{mydarkblue!20}66.10 & \cellcolor{mydarkblue!14}57.27 & \cellcolor{mydarkblue!23}71.90 & \cellcolor{mydarkblue!23}86.59 & \cellcolor{mydarkblue!22}78.08 & \cellcolor{mydarkblue!24}90.34 & \cellcolor{mydarkblue!24}90.93 & \cellcolor{mydarkblue!23}83.76 & \cellcolor{mydarkblue!24}93.70 & \cellcolor{mydarkblue!21}78.34 & \cellcolor{mydarkblue!18}72.45 & \cellcolor{mydarkblue!18}82.16 & \cellcolor{mydarkblue!22}79.30 \\
    \hline
    \href{https://huggingface.co/meta-llama/Llama-3.1-70B-Instruct}{\color{mydarkblue} \textbf{\footnotesize Llama-3.1-70B-Instruct}} & 70B & \cellcolor{mydarkblue!7}58.88 & \cellcolor{mydarkblue!11}54.94 & \cellcolor{mydarkblue!18}62.18 & \cellcolor{mydarkblue!11}69.85 & \cellcolor{mydarkblue!5}58.36 & \cellcolor{mydarkblue!15}76.36 & \cellcolor{mydarkblue!11}63.93 & \cellcolor{mydarkblue!4}53.06 & \cellcolor{mydarkblue!6}70.71 & \cellcolor{mydarkblue!17}71.94 & \cellcolor{mydarkblue!3}55.76 & \cellcolor{mydarkblue!15}79.45 & \cellcolor{mydarkblue!9}64.62 \\
    \ + CoT {\scriptsize \color{mydarkblue} (NeurIPS'22)} & 70B & \cellcolor{mydarkblue!16}63.70 & \cellcolor{mydarkblue!1}44.75 & \cellcolor{mydarkblue!24}72.97 & \cellcolor{mydarkblue!20}82.47 & \cellcolor{mydarkblue!15}70.09 & \cellcolor{mydarkblue!22}87.61 & \cellcolor{mydarkblue!21}85.22 & \cellcolor{mydarkblue!13}66.97 & \cellcolor{mydarkblue!22}90.48 & \cellcolor{mydarkblue!21}79.32 & \cellcolor{mydarkblue!9}62.36 & \cellcolor{mydarkblue!21}85.74 & \cellcolor{mydarkblue!18}74.31 \\
    \ +  ARG {\scriptsize \color{mydarkblue} (AAAI'24)} & 70B & \cellcolor{mydarkblue!13}62.32 & \cellcolor{mydarkblue!1}44.33 & \cellcolor{mydarkblue!23}71.53 & \cellcolor{mydarkblue!22}84.62 & \cellcolor{mydarkblue!19}74.39 & \cellcolor{mydarkblue!23}89.01 & \cellcolor{mydarkblue!21}84.67 & \cellcolor{mydarkblue!14}69.03 & \cellcolor{mydarkblue!21}89.81 & \cellcolor{mydarkblue!20}78.13 & \cellcolor{mydarkblue!4}57.34 & \cellcolor{mydarkblue!21}85.30 & \cellcolor{mydarkblue!18}74.21 \\
    \ + \textsc{GenFend} {\scriptsize \color{mydarkblue} (CIKM'24)} & 70B & \cellcolor{mydarkblue!12}61.50 & \cellcolor{mydarkblue!0}40.36 & \cellcolor{mydarkblue!23}71.58 & \cellcolor{mydarkblue!20}82.73 & \cellcolor{mydarkblue!17}71.33 & \cellcolor{mydarkblue!22}87.65 & \cellcolor{mydarkblue!21}83.42 & \cellcolor{mydarkblue!9}61.56 & \cellcolor{mydarkblue!21}89.43 & \cellcolor{mydarkblue!17}71.94 & \cellcolor{mydarkblue!1}53.62 & \cellcolor{mydarkblue!16}79.88 & \cellcolor{mydarkblue!15}71.25 \\
    \ + DMR {\scriptsize \color{mydarkblue} (ACL'25)} & 70B & \cellcolor{mydarkblue!16}64.15 & \cellcolor{mydarkblue!10}53.26 & \cellcolor{mydarkblue!23}70.92 & \cellcolor{mydarkblue!18}78.95 & \cellcolor{mydarkblue!14}68.47 & \cellcolor{mydarkblue!20}84.20 & \cellcolor{mydarkblue!20}81.66 & \cellcolor{mydarkblue!13}67.32 & \cellcolor{mydarkblue!19}87.26 & \cellcolor{mydarkblue!19}75.77 & \cellcolor{mydarkblue!9}63.06 & \cellcolor{mydarkblue!18}81.97 & \cellcolor{mydarkblue!17}73.08 \\
    \ + PCoT {\scriptsize \color{mydarkblue} (ACL'25)} & 70B & \cellcolor{mydarkblue!13}62.05 & \cellcolor{mydarkblue!7}50.39 & \cellcolor{mydarkblue!22}69.27 & \cellcolor{mydarkblue!16}77.41 & \cellcolor{mydarkblue!13}67.25 & \cellcolor{mydarkblue!19}82.75 & \cellcolor{mydarkblue!22}85.87 & \cellcolor{mydarkblue!19}77.18 & \cellcolor{mydarkblue!21}89.77 & \cellcolor{mydarkblue!22}80.85 & \cellcolor{mydarkblue!16}70.59 & \cellcolor{mydarkblue!21}85.80 & \cellcolor{mydarkblue!18}74.93 \\
    \hline
    \href{https://huggingface.co/deepseek-ai/DeepSeek-R1-Distill-Qwen-32B}{\color{mydarkblue} \textbf{\footnotesize R1-Distill-Qwen-32B}} & 671/32B & \cellcolor{mydarkblue!19}65.75 & \cellcolor{mydarkblue!10}54.02 & \cellcolor{mydarkblue!24}72.70 & \cellcolor{mydarkblue!25}88.65 & \cellcolor{mydarkblue!24}80.29 & \cellcolor{mydarkblue!25}92.03 & \cellcolor{mydarkblue!21}84.91 & \cellcolor{mydarkblue!17}73.10 & \cellcolor{mydarkblue!21}89.52 & \cellcolor{mydarkblue!17}72.28 & \cellcolor{mydarkblue!1}53.39 & \cellcolor{mydarkblue!16}80.27 & \cellcolor{mydarkblue!19}75.58 \\
    \ + CoT {\scriptsize \color{mydarkblue} (NeurIPS'22)} & 671/32B & \cellcolor{mydarkblue!21}66.52 & \cellcolor{mydarkblue!13}57.12 & \cellcolor{mydarkblue!24}72.54 & \cellcolor{mydarkblue!25}89.08 & \cellcolor{mydarkblue!25}81.24 & \cellcolor{mydarkblue!25}92.30 & \cellcolor{mydarkblue!20}83.01 & \cellcolor{mydarkblue!15}70.23 & \cellcolor{mydarkblue!20}88.11 & \cellcolor{mydarkblue!18}73.88 & \cellcolor{mydarkblue!5}58.37 & \cellcolor{mydarkblue!17}80.97 & \cellcolor{mydarkblue!19}76.11 \\
    \ +  ARG {\scriptsize \color{mydarkblue} (AAAI'24)} & 671/32B & \cellcolor{mydarkblue!18}65.17 & \cellcolor{mydarkblue!16}59.91 & \cellcolor{mydarkblue!22}69.21 & \cellcolor{mydarkblue!11}70.18 & \cellcolor{mydarkblue!9}62.56 & \cellcolor{mydarkblue!14}75.23 & \cellcolor{mydarkblue!16}73.74 & \cellcolor{mydarkblue!9}60.36 & \cellcolor{mydarkblue!14}80.37 & \cellcolor{mydarkblue!14}67.47 & \cellcolor{mydarkblue!3}55.22 & \cellcolor{mydarkblue!10}74.46 & \cellcolor{mydarkblue!11}67.82 \\
    \ + \textsc{GenFend} {\scriptsize \color{mydarkblue} (CIKM'24)} & 671/32B & \cellcolor{mydarkblue!14}62.72 & \cellcolor{mydarkblue!0}43.37 & \cellcolor{mydarkblue!24}72.21 & \cellcolor{mydarkblue!23}86.34 & \cellcolor{mydarkblue!20}75.34 & \cellcolor{mydarkblue!24}90.55 & \cellcolor{mydarkblue!22}87.37 & \cellcolor{mydarkblue!17}74.07 & \cellcolor{mydarkblue!22}91.65 & \cellcolor{mydarkblue!21}78.76 & \cellcolor{mydarkblue!8}61.14 & \cellcolor{mydarkblue!21}85.38 & \cellcolor{mydarkblue!19}75.74 \\
    \ + DMR {\scriptsize \color{mydarkblue} (ACL'25)} & 671/32B & \cellcolor{mydarkblue!19}65.37 & \cellcolor{mydarkblue!14}58.09 & \cellcolor{mydarkblue!23}70.50 & \cellcolor{mydarkblue!23}86.76 & \cellcolor{mydarkblue!22}77.80 & \cellcolor{mydarkblue!24}90.57 & \cellcolor{mydarkblue!19}81.11 & \cellcolor{mydarkblue!14}69.42 & \cellcolor{mydarkblue!18}86.33 & \cellcolor{mydarkblue!18}73.11 & \cellcolor{mydarkblue!6}59.36 & \cellcolor{mydarkblue!16}79.91 & \cellcolor{mydarkblue!18}74.86 \\
    \ + PCoT {\scriptsize \color{mydarkblue} (ACL'25)} & 671/32B & \cellcolor{mydarkblue!18}64.72 & \cellcolor{mydarkblue!5}48.02 & \cellcolor{mydarkblue!24}73.30 & \cellcolor{mydarkblue!24}87.54 & \cellcolor{mydarkblue!21}77.51 & \cellcolor{mydarkblue!25}91.38 & \cellcolor{mydarkblue!24}90.38 & \cellcolor{mydarkblue!21}80.87 & \cellcolor{mydarkblue!24}93.57 & \cellcolor{mydarkblue!24}82.79 & \cellcolor{mydarkblue!14}67.88 & \cellcolor{mydarkblue!24}88.25 & \cellcolor{mydarkblue!22}78.85 \\
    \hline
    \href{https://huggingface.co/Qwen/Qwen3-32B}{\color{mydarkblue} \textbf{\footnotesize Qwen3-32B}} & 32B & \cellcolor{mydarkblue!23}67.70 & \cellcolor{mydarkblue!16}60.24 & \cellcolor{mydarkblue!24}72.80 & \cellcolor{mydarkblue!24}87.28 & \cellcolor{mydarkblue!23}79.03 & \cellcolor{mydarkblue!24}90.87 & \cellcolor{mydarkblue!23}89.72 & \cellcolor{mydarkblue!22}81.74 & \cellcolor{mydarkblue!23}92.85 & \cellcolor{mydarkblue!23}81.54 & \cellcolor{mydarkblue!18}72.65 & \cellcolor{mydarkblue!22}86.07 & \cellcolor{mydarkblue!23}80.21 \\
    \ + CoT {\scriptsize \color{mydarkblue} (NeurIPS'22)} & 32B & \cellcolor{mydarkblue!14}62.53 & \cellcolor{mydarkblue!12}55.19 & \cellcolor{mydarkblue!21}67.81 & \cellcolor{mydarkblue!23}86.96 & \cellcolor{mydarkblue!23}78.90 & \cellcolor{mydarkblue!24}90.57 & \cellcolor{mydarkblue!22}86.57 & \cellcolor{mydarkblue!19}76.61 & \cellcolor{mydarkblue!22}90.58 & \cellcolor{mydarkblue!19}76.11 & \cellcolor{mydarkblue!10}63.78 & \cellcolor{mydarkblue!18}82.18 & \cellcolor{mydarkblue!20}76.48 \\
    \ +  ARG {\scriptsize \color{mydarkblue} (AAAI'24)} & 32B & \cellcolor{mydarkblue!8}59.43 & \cellcolor{mydarkblue!0}41.46 & \cellcolor{mydarkblue!22}68.96 & \cellcolor{mydarkblue!17}77.84 & \cellcolor{mydarkblue!2}55.20 & \cellcolor{mydarkblue!20}85.28 & \cellcolor{mydarkblue!22}86.57 & \cellcolor{mydarkblue!19}76.61 & \cellcolor{mydarkblue!22}90.58 & \cellcolor{mydarkblue!19}76.11 & \cellcolor{mydarkblue!10}63.78 & \cellcolor{mydarkblue!18}82.18 & \cellcolor{mydarkblue!16}72.00 \\
    \ + \textsc{GenFend} {\scriptsize \color{mydarkblue} (CIKM'24)} & 32B & \cellcolor{mydarkblue!12}61.78 & \cellcolor{mydarkblue!11}54.34 & \cellcolor{mydarkblue!20}67.14 & \cellcolor{mydarkblue!24}87.05 & \cellcolor{mydarkblue!23}79.26 & \cellcolor{mydarkblue!24}90.59 & \cellcolor{mydarkblue!21}85.12 & \cellcolor{mydarkblue!18}74.85 & \cellcolor{mydarkblue!21}89.43 & \cellcolor{mydarkblue!18}74.58 & \cellcolor{mydarkblue!8}61.61 & \cellcolor{mydarkblue!17}80.99 & \cellcolor{mydarkblue!19}75.56 \\
    \ + DMR {\scriptsize \color{mydarkblue} (ACL'25)} & 32B & \cellcolor{mydarkblue!14}62.75 & \cellcolor{mydarkblue!14}57.64 & \cellcolor{mydarkblue!20}66.75 & \cellcolor{mydarkblue!24}87.54 & \cellcolor{mydarkblue!24}79.94 & \cellcolor{mydarkblue!24}90.96 & \cellcolor{mydarkblue!17}75.20 & \cellcolor{mydarkblue!11}64.71 & \cellcolor{mydarkblue!14}80.88 & \cellcolor{mydarkblue!18}74.02 & \cellcolor{mydarkblue!11}64.44 & \cellcolor{mydarkblue!16}79.53 & \cellcolor{mydarkblue!17}73.70 \\
    \ + PCoT {\scriptsize \color{mydarkblue} (ACL'25)} & 32B & \cellcolor{mydarkblue!22}67.12 & \cellcolor{mydarkblue!11}54.98 & \cellcolor{mydarkblue!25}74.10 & \cellcolor{mydarkblue!23}86.76 & \cellcolor{mydarkblue!22}77.74 & \cellcolor{mydarkblue!24}90.58 & \cellcolor{mydarkblue!24}89.77 & \cellcolor{mydarkblue!21}80.60 & \cellcolor{mydarkblue!24}93.06 & \cellcolor{mydarkblue!24}83.28 & \cellcolor{mydarkblue!21}76.00 & \cellcolor{mydarkblue!23}87.17 & \cellcolor{mydarkblue!23}80.10 \\
    \hline
    \href{https://huggingface.co/Qwen/Qwen3-30B-A3B-Instruct-2507}{\color{mydarkblue} \textbf{\footnotesize Qwen3-30B-A3B-Instr.}} & 30B & \cellcolor{mydarkblue!22}67.03 & \cellcolor{mydarkblue!24}68.24 & \cellcolor{mydarkblue!19}65.71 & \cellcolor{mydarkblue!18}78.87 & \cellcolor{mydarkblue!17}72.05 & \cellcolor{mydarkblue!19}83.01 & \cellcolor{mydarkblue!23}88.03 & \cellcolor{mydarkblue!21}80.58 & \cellcolor{mydarkblue!22}91.34 & \cellcolor{mydarkblue!20}76.67 & \cellcolor{mydarkblue!17}71.34 & \cellcolor{mydarkblue!16}80.33 & \cellcolor{mydarkblue!20}76.93 \\
    \ + CoT {\scriptsize \color{mydarkblue} (NeurIPS'22)} & 30B & \cellcolor{mydarkblue!18}64.75 & \cellcolor{mydarkblue!22}66.28 & \cellcolor{mydarkblue!17}63.07 & \cellcolor{mydarkblue!21}83.76 & \cellcolor{mydarkblue!22}78.00 & \cellcolor{mydarkblue!22}87.13 & \cellcolor{mydarkblue!20}81.31 & \cellcolor{mydarkblue!16}72.75 & \cellcolor{mydarkblue!18}85.78 & \cellcolor{mydarkblue!20}76.74 & \cellcolor{mydarkblue!17}72.21 & \cellcolor{mydarkblue!16}80.00 & \cellcolor{mydarkblue!19}75.98 \\
    \ +  ARG {\scriptsize \color{mydarkblue} (AAAI'24)} & 30B & \cellcolor{mydarkblue!16}63.92 & \cellcolor{mydarkblue!12}55.37 & \cellcolor{mydarkblue!22}69.73 & \cellcolor{mydarkblue!20}81.87 & \cellcolor{mydarkblue!18}72.70 & \cellcolor{mydarkblue!21}86.43 & \cellcolor{mydarkblue!23}88.03 & \cellcolor{mydarkblue!20}78.83 & \cellcolor{mydarkblue!22}91.65 & \cellcolor{mydarkblue!19}76.04 & \cellcolor{mydarkblue!10}63.33 & \cellcolor{mydarkblue!18}82.21 & \cellcolor{mydarkblue!19}75.84 \\
    \ + \textsc{GenFend} {\scriptsize \color{mydarkblue} (CIKM'24)} & 30B & \cellcolor{mydarkblue!22}67.32 & \cellcolor{mydarkblue!18}61.39 & \cellcolor{mydarkblue!23}71.68 & \cellcolor{mydarkblue!21}83.33 & \cellcolor{mydarkblue!20}75.32 & \cellcolor{mydarkblue!22}87.42 & \cellcolor{mydarkblue!23}89.33 & \cellcolor{mydarkblue!21}80.72 & \cellcolor{mydarkblue!23}92.62 & \cellcolor{mydarkblue!20}76.67 & \cellcolor{mydarkblue!17}72.43 & \cellcolor{mydarkblue!16}79.78 & \cellcolor{mydarkblue!24}81.17 \\
    \ + DMR {\scriptsize \color{mydarkblue} (ACL'25)} & 30B & \cellcolor{mydarkblue!16}63.60 & \cellcolor{mydarkblue!22}65.74 & \cellcolor{mydarkblue!16}61.17 & \cellcolor{mydarkblue!16}76.29 & \cellcolor{mydarkblue!16}71.19 & \cellcolor{mydarkblue!17}79.85 & \cellcolor{mydarkblue!14}69.54 & \cellcolor{mydarkblue!10}62.84 & \cellcolor{mydarkblue!9}74.19 & \cellcolor{mydarkblue!16}71.52 & \cellcolor{mydarkblue!14}68.37 & \cellcolor{mydarkblue!10}74.10 & \cellcolor{mydarkblue!14}69.87 \\
    \ + PCoT {\scriptsize \color{mydarkblue} (ACL'25)} & 30B & \cellcolor{mydarkblue!25}68.73 & \cellcolor{mydarkblue!25}69.07 & \cellcolor{mydarkblue!21}68.37 & \cellcolor{mydarkblue!19}81.36 & \cellcolor{mydarkblue!19}73.63 & \cellcolor{mydarkblue!21}85.58 & \cellcolor{mydarkblue!24}90.33 & \cellcolor{mydarkblue!24}84.24 & \cellcolor{mydarkblue!23}93.02 & \cellcolor{mydarkblue!25}85.03 & \cellcolor{mydarkblue!24}80.79 & \cellcolor{mydarkblue!23}87.74 & \cellcolor{mydarkblue!23}80.66 \\
    \hline
    \href{https://huggingface.co/openai/gpt-oss-20b}{\color{mydarkblue} \footnotesize \textbf{gpt-oss-20b} (high effort)} & 20B & \cellcolor{mydarkblue!2}56.20 & \cellcolor{mydarkblue!23}67.14 & \cellcolor{mydarkblue!1}34.33 & \cellcolor{mydarkblue!3}58.76 & \cellcolor{mydarkblue!5}58.69 & \cellcolor{mydarkblue!4}58.83 & \cellcolor{mydarkblue!13}66.33 & \cellcolor{mydarkblue!8}59.90 & \cellcolor{mydarkblue!6}70.98 & \cellcolor{mydarkblue!15}68.38 & \cellcolor{mydarkblue!9}62.91 & \cellcolor{mydarkblue!9}72.45 & \cellcolor{mydarkblue!5}61.24 \\
    \ + CoT {\scriptsize \color{mydarkblue} (NeurIPS'22)} & 20B & \cellcolor{mydarkblue!0}54.87 & \cellcolor{mydarkblue!23}66.75 & \cellcolor{mydarkblue!0}29.79 & \cellcolor{mydarkblue!0}55.07 & \cellcolor{mydarkblue!4}57.03 & \cellcolor{mydarkblue!0}52.93 & \cellcolor{mydarkblue!9}59.67 & \cellcolor{mydarkblue!6}55.84 & \cellcolor{mydarkblue!0}62.89 & \cellcolor{mydarkblue!11}64.14 & \cellcolor{mydarkblue!8}60.84 & \cellcolor{mydarkblue!3}66.92 & \cellcolor{mydarkblue!2}57.23 \\
    \ +  ARG {\scriptsize \color{mydarkblue} (AAAI'24)} & 20B & \cellcolor{mydarkblue!0}52.22 & \cellcolor{mydarkblue!23}66.86 & \cellcolor{mydarkblue!0}14.42 & \cellcolor{mydarkblue!0}41.75 & \cellcolor{mydarkblue!0}52.12 & \cellcolor{mydarkblue!0}25.66 & \cellcolor{mydarkblue!0}40.68 & \cellcolor{mydarkblue!0}46.52 & \cellcolor{mydarkblue!0}33.41 & \cellcolor{mydarkblue!0}44.29 & \cellcolor{mydarkblue!1}52.27 & \cellcolor{mydarkblue!0}33.11 & \cellcolor{mydarkblue!0}41.94 \\
    \ + \textsc{GenFend} {\scriptsize \color{mydarkblue} (CIKM'24)} & 20B & \cellcolor{mydarkblue!0}48.39 & \cellcolor{mydarkblue!15}57.89 & \cellcolor{mydarkblue!0}33.33 & \cellcolor{mydarkblue!0}52.75 & \cellcolor{mydarkblue!0}51.75 & \cellcolor{mydarkblue!0}53.70 & \cellcolor{mydarkblue!8}56.41 & \cellcolor{mydarkblue!0}46.82 & \cellcolor{mydarkblue!0}63.07 & \cellcolor{mydarkblue!8}58.98 & \cellcolor{mydarkblue!0}51.68 & \cellcolor{mydarkblue!1}64.37 & \cellcolor{mydarkblue!0}53.26 \\
    \ + DMR {\scriptsize \color{mydarkblue} (ACL'25)} & 20B & \cellcolor{mydarkblue!1}55.65 & \cellcolor{mydarkblue!20}63.80 & \cellcolor{mydarkblue!6}42.77 & \cellcolor{mydarkblue!4}61.17 & \cellcolor{mydarkblue!5}57.84 & \cellcolor{mydarkblue!7}64.01 & \cellcolor{mydarkblue!11}63.78 & \cellcolor{mydarkblue!5}54.95 & \cellcolor{mydarkblue!5}69.71 & \cellcolor{mydarkblue!13}65.46 & \cellcolor{mydarkblue!4}57.02 & \cellcolor{mydarkblue!7}71.13 & \cellcolor{mydarkblue!5}60.61 \\
    \ + PCoT {\scriptsize \color{mydarkblue} (ACL'25)} & 20B & \cellcolor{mydarkblue!15}63.48 & \cellcolor{mydarkblue!22}66.16 & \cellcolor{mydarkblue!21}60.33 & \cellcolor{mydarkblue!15}75.00 & \cellcolor{mydarkblue!15}69.59 & \cellcolor{mydarkblue!16}78.77 & \cellcolor{mydarkblue!22}86.17 & \cellcolor{mydarkblue!20}78.03 & \cellcolor{mydarkblue!21}89.91 & \cellcolor{mydarkblue!21}77.79 & \cellcolor{mydarkblue!16}70.10 & \cellcolor{mydarkblue!18}82.33 & \cellcolor{mydarkblue!18}74.81 \\
    \hline
    \specialrule{0em}{0.5pt}{0.5pt}
    \hline
    \multicolumn{15}{c}{\textbf{$<$ 10B parameters}} \\
    \rowcolor{lightgrayv} \textbf{\baby-4B} {\scriptsize \color{mydarkblue}\ (Ours)} & 4B & \cellcolor{mydarkblue!23}\textbf{67.98} & \cellcolor{mydarkblue!21}\textbf{65.26} & \cellcolor{mydarkblue!22}\textbf{70.29} & \cellcolor{mydarkblue!24}\textbf{87.11} & \cellcolor{mydarkblue!23}\textbf{78.57} & \cellcolor{mydarkblue!24}\textbf{90.79} & \cellcolor{mydarkblue!25}\textbf{92.28} & \cellcolor{mydarkblue!25}\textbf{86.47} & \cellcolor{mydarkblue!25}\textbf{94.60} & \cellcolor{mydarkblue!25}\textbf{85.03} & \cellcolor{mydarkblue!23}\textbf{78.13} & \cellcolor{mydarkblue!24}\textbf{88.62} & \cellcolor{mydarkblue!25}\textbf{82.09} \\
    \rowcolor{lightgrayv} \textbf{\baby-2B} {\scriptsize \color{mydarkblue}\ (Ours)} & 2B & \cellcolor{mydarkblue!22}\textbf{67.12} & \cellcolor{mydarkblue!21}\textbf{65.09} & \cellcolor{mydarkblue!21}\textbf{68.93} & \cellcolor{mydarkblue!19}\textbf{81.53} & \cellcolor{mydarkblue!12}\textbf{66.46} & \cellcolor{mydarkblue!21}\textbf{87.26} & \cellcolor{mydarkblue!23}\textbf{89.58} & \cellcolor{mydarkblue!22}\textbf{82.16} & \cellcolor{mydarkblue!23}\textbf{92.64} & \cellcolor{mydarkblue!21}\textbf{78.34} & \cellcolor{mydarkblue!16}\textbf{70.12} & \cellcolor{mydarkblue!19}\textbf{83.01} & \cellcolor{mydarkblue!21}\textbf{77.69} \\
    \rowcolor{lightgrayv} \textbf{\baby-1.5B} {\scriptsize \color{mydarkblue}\ (Ours)} & 1.5B & \cellcolor{mydarkblue!24}\textbf{68.50} & \cellcolor{mydarkblue!25}\textbf{69.01} & \cellcolor{mydarkblue!20}\textbf{67.97} & \cellcolor{mydarkblue!22}\textbf{84.97} & \cellcolor{mydarkblue!20}\textbf{75.59} & \cellcolor{mydarkblue!23}\textbf{89.14} & \cellcolor{mydarkblue!24}\textbf{90.63} & \cellcolor{mydarkblue!23}\textbf{83.17} & \cellcolor{mydarkblue!24}\textbf{93.51} & \cellcolor{mydarkblue!24}\textbf{83.15} & \cellcolor{mydarkblue!18}\textbf{72.31} & \cellcolor{mydarkblue!23}\textbf{87.89} & \cellcolor{mydarkblue!23}\textbf{80.49} \\
    \bottomrule
  \end{tabular} }
\end{table*}

\begin{figure*}
  \centering
  \includegraphics[width=0.98\textwidth]{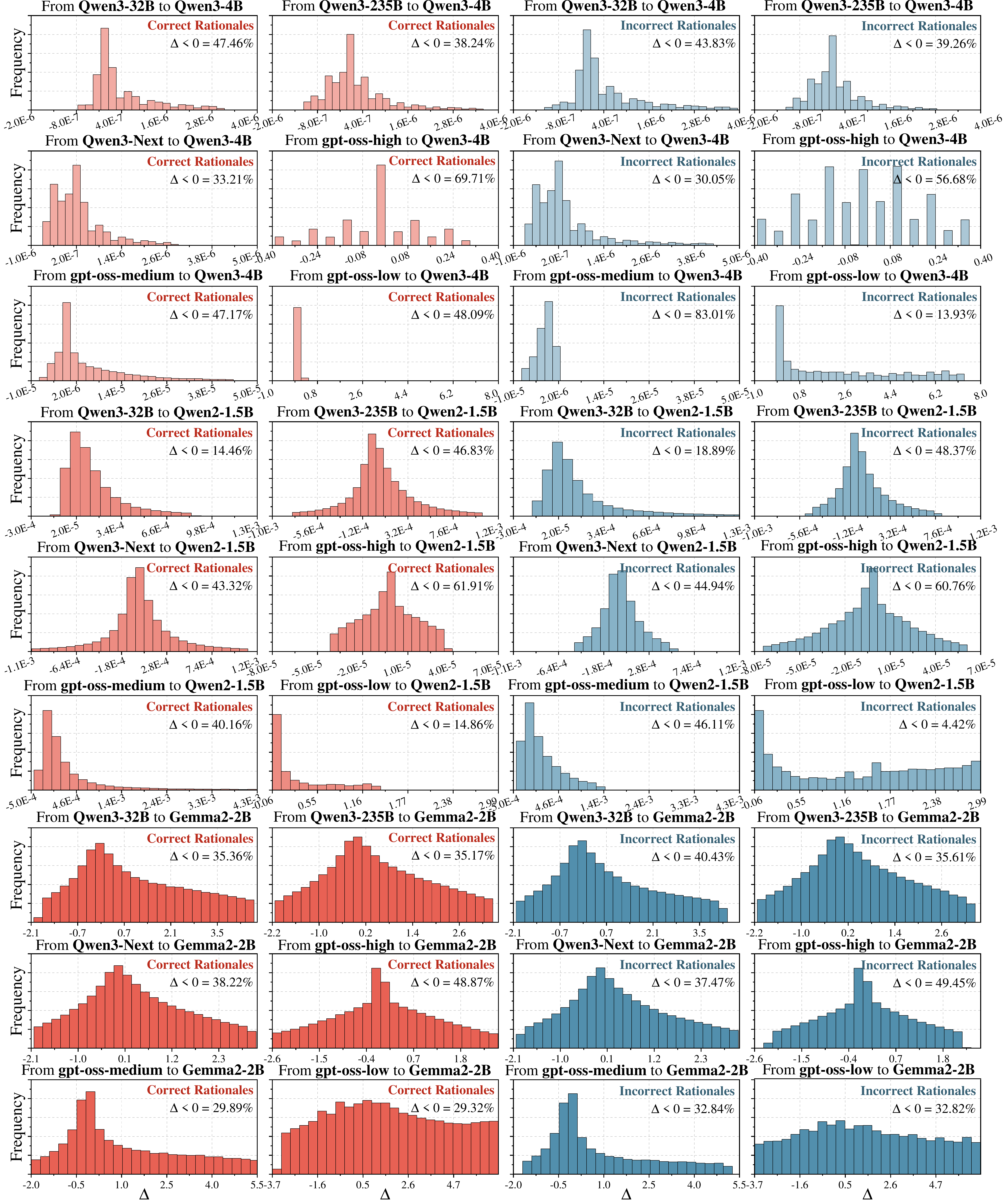}
  \vspace{-2pt}
  \caption{Distribution of $\Delta$ for correct / incorrect rationales.}
  \label{all_deltas}
  \vspace{-3pt}
\end{figure*}